# Learning Geometrically-Constrained Hidden Markov Models for Robot Navigation: Bridging the Topological-Geometrical Gap

**Hagit Shatkay**                                                    HAGIT.SHATKAY@CELERA.COM
*Informatics Research Group,*
*Celera Genomics, Rockville, MD 20850*

**Leslie Pack Kaelbling**                                            LPK@AI.MIT.EDU
*Artificial Intelligence Laboratory*
*Massachusetts Institute of Technology, Cambridge, MA 02139*

> *You will come to a place where the streets are not marked.*
> *Some windows are lighted but mostly they're darked.*
> *A place you could sprain both your elbow and chin!*
> *Do you dare to stay out? Do you dare to go in?...*
> *And if you go in, should you turn left or right...*
> *or right-and-three-quarters? or, maybe, not quite?...*
> *Simple it's not, I'm afraid you will find,*
> *for a mind-maker-upper to make up his mind.*
>
> *Oh, the Places You'll Go,* Dr. Seuss.

## Abstract

Hidden Markov models (HMMs) and partially observable Markov decision processes (POMDPs) provide useful tools for modeling dynamical systems. They are particularly useful for representing the topology of environments such as road networks and office buildings, which are typical for robot navigation and planning. The work presented here describes a formal framework for incorporating readily available odometric information and geometrical constraints into both the models and the algorithm that learns them. By taking advantage of such information, learning HMMs/POMDPs can be made to generate better solutions and require fewer iterations, while being robust in the face of data reduction. Experimental results, obtained from both simulated and real robot data, demonstrate the effectiveness of the approach.

## 1 Introduction

This work is concerned with robots that need to perform tasks in *structured environments*. A robot moving in the environment suffers from two main limitations: its noisy sensors prevent it from confidently knowing where it is, while its noisy effectors prevent it from knowing with certainty where its actions will take it. We concentrate here on *structured* environments, which can in turn be characterized by two main properties: such environments consist of vast *uneventful* and uninteresting areas, and are interspersed with relatively *few interesting positions or situations.* Consider for instance a robot delivering a bagel in an office building. The interesting situations are the doors and the intersections in the building hallways, as well as the various positions where the bagel might be with respect to the robot's arm (e.g., the robot is holding the bagel, puts it down, etc.) Most other aspects of the environment, such as the desk positions in the offices, are inconsequential for the bagel delivery task.

A natural way to represent the *combination* of such an *environment* and the robot's *interactions* with it, is as a probabilistic automaton, in which states represent interesting situations, and edges between states represent the *actions* leading from one situation to another. Probability distributions over the transitions and over the possible observations the robot may perceive at each situation model the robot's noisy effectors and sensors, respectively.

Such models are formally known as POMDP (partially observable Markov decision process) models, and have been proven useful for robot planning and acting under the inherent world uncertainty (Simmons & Koenig, 1995; Nourbakhsh, Powers, & Birchfield, 1995; Cassandra, Kaelbling, & Kurien, 1996).

Despite much work on using such models, the task of learning them directly and automatically from the data has not been widely addressed. Research concerning this immediate topic to date consists mostly of the work done by Simmons and Koenig (1996b). The assumption underlying their work was that a human provides a rather accurate topological model of the states and their connections, and the exact probability distributions are then learned on top of this model, using a version of the Baum-Welch algorithm (Rabiner, 1989). Another interesting approach to the acquisition of topological models is that of Thrun and Bücken (1996a,1996b; Thrun, 1999), who focused on extracting deterministic topological maps from previously acquired geometrical-grid-based maps, where the latter were learned directly from the data. Further discussion of related research on both the geometrical and the topological approaches, in their probabilistic and deterministic versions, is given in the next section.

The work reported here is the first successful attempt we are aware of to learn purely *probabilistic-topological* models, directly and completely from recorded data, without using previous human-provided or grid-based models. It is based on using *weak geometric* information, recorded by the robot, to help learn the *topology* of the environment, and represent it as a probabilistic model. Therefore, it directly bridges the historically perceived gap between topological and geometrical information, and addresses the claim presented in Thrun's work (1999) that the main shortcoming of the topological approach is its failure to utilize the inherent geometry of the learnt environment.

Most robots are equipped with wheel encoders that enable an odometer to record the change in the robot's position as it moves through the environment. This data is typically very noisy and inaccurate. The floors in the environment are rarely smooth, the wheels of the robot are not always aligned and neither are the motors, the mechanics is imperfect, resulting in slippage and drift. All these effects accumulate, and if we were to mark the initial position of the robot, and try to estimate its current position based on summing a long sequence of odometric recordings, the resulting estimate will be incorrect. That is, the raw recorded odometric information is not an effective tool, in and of itself, for determining the absolute location of the robot in the environment.

While our approach is *not* aimed at determining absolute locations, the idea underlying it is that this weak odometric information, despite its noise and inaccuracy, still provides geometrical cues that can help to distinguish between different states, as well as to identify revisitation of the same state. Hence, such information enhances the ability to learn *topological* models. However, the use of geometrical information requires careful treatment of geometrical constraints and directional data. We demonstrate how the existing models and algorithms can be extended to take advantage of the noisy odometric data and the geometrical constraints. The geometrical information is directly incorporated into the probabilistic topological framework, producing a significant improvement over the standard Baum-Welch algorithm, without the need for human-provided model.

The rest of this paper is organized as follows: Section 2 provides a survey of previous work in the area of learning maps for robot navigation, and briefly refers to earlier work on learning automata; Section 3 presents the formal framework for this work; Section 4 presents the main aspects of our iterative learning algorithm, while Section 5 describes the strategies for selecting the initial point from which the iterative process begins; Section 6 presents experimental results obtained from both simulated and real robot data in traditionally hard-to-learn environments. The experiments demonstrate that our algorithm indeed converges to better models with fewer iterations than the standard Baum-Welch method, and is robust in the face of data reduction.

## 2 Approaches to Learning Maps and Models

The work presented here lies in the intersection between the theoretical area of learning computational models—in particular, learning automata from data sequences—and the applied area of map acquisition for robot navigation. We concentrate here on surveying the work in the latter area, pointing out the distinction between our approach and its predecessors. We briefly review some results from automata and computational learning theory. A more comprehensive review of theoretical results is given by Shatkay (1999).

### 2.1 Modeling Environments for Robot Navigation

In the context of maps and models for robot navigation, a distinction is usually made between two principal kinds of maps: *geometric* and *topological*. Geometric maps describe the environment as a collection of *objects* or *occupied positions* in space, and the *geometric* relationships among them. The *topological* framework is less concerned with the geometrical positions, and models the world as a collection of *states* and their *connectivity*, that is, which states are reachable from each of the other states and what actions lead from one state to the next.

We draw an additional distinction, between world-centric[1] *maps* that provide an "objective" description of the environment independent of the agent using the map, and robot-centric *models* which capture the *interaction* of a particular "subjective" agent with the environment. When learning a *map*, the agent needs to take into account its own noisy sensors and actuators and try to obtain an objectively correct map that other agents could use as well. Similarly, other agents using the map need to compensate for their own limitations in order to assess their position according to the map. When learning a *model* that captures *interaction*, the agent acquiring the model is the one who is also using it. Hence, the noisy sensors and actuators specific to the agent are reflected in the model. A different model is likely to be needed by different agents. Most of the related work described below, especially within the geometrical framework, is centered around learning objective maps of the world rather than agent-specific models. We shall point out in this survey the work that is concerned with the latter kind of models.

Our work focuses on acquiring *purely topological models*, and is less concerned with learning geometrical relationships between locations or objects, or objective maps, although geometrical relationships do serve as an aid in our acquisition process. The concept of a *state* used in this topological framework is more general than the concept of a *geometrical location*, since a state can include information such as the battery level, the arm position etc. Such information, which is of great importance for planning, is non-geometrical in nature and therefore cannot be readily captured in a purely geometrical framework. The following sections provide a survey of work done both within the geometrical framework and within the topological framework, as well as combinations of the two approaches.

---

1. We thank Sebastian Thrun for the terminology.

## 2.2 Geometric Maps

Geometric maps provide a description of the environment in terms of the objects placed in it and their positions. For example, *grid-based* maps are an instance of the geometric approach. In a grid-based map, the environment is modeled as a grid (an array), where each position in the grid can be either vacant or occupied by some object (binary values placed in the array). This approach can be further refined to reflect uncertainty about the world, by having grid cells contain occupancy probabilities rather than just binary values. A lot of work has been done on learning such grid-based maps for robot navigation through the use of sonar readings and their interpretation, by Moravec and Elfes and others (Moravec & Elfes, 1985; Moravec, 1988; Elfes, 1989; Asada, 1991).

An underlying assumption when learning such maps is that the robot can tell (or find out) where it is on the grid when it obtains a sonar reading indicating an object, and therefore can place the object correctly on the grid. A similar localization assumption, requiring the robot to identify its geometrical location, underlies other geometric mapping techniques by Leonard et al. (1991), Smith et al. (1991), Thrun et al. (1998b) and Dissanayake et al. (2001), even when an explicit grid is not part of the model. Explicit localization can be hard to satisfy. Leonard et al. (1991) and Smith et al. (1991) address this issue through the use of geometrical beacons to estimate the location of the robot. In what is known as the *Kalman filter* method, a Gaussian probability distribution is used to model the robot's possible current location, based on observations collected up to the current point, (without allowing the refinement of previous position estimates based on later observations). Research in this area has recently been extended in two directions: Leonard and Feder (2000) partition the task of learning one large map into learning multiple smaller map-sections, thus addressing the issue of computational efficiency. Dissanayake et al. (2001) conduct a theoretical study of the approach and show its convergence properties. The latter may lead to computational efficiency by identifying the cases for which a steady-state solution can be readily obtained, accordingly bounding the number of steps required by the algorithms to reach a useful solution in these cases.

Work by Thrun et al. (1998a) uses a similar probabilistic approach for obtaining grid-based maps. This work is refined (Thrun et al., 1998b) to first learn the location of significant landmarks in the environment and then fill in the details of the complete geometrical grid, based on laser range scans. The latter work extends the approach of Smith et al. , by using observations obtained both *before* and *after* a location has been visited, in order to derive a probability distribution over possible locations. To achieve this, the authors use a *forward-backward* procedure similar to the one used in the Baum-Welch algorithm (Rabiner, 1989), in order to determine possible locations from observed data. The approach resembles ours both in the use of the *forward-backward* estimation procedure, and in its probabilistic basis, aiming at obtaining a maximum likelihood map of the environment. It still significantly differs from ours both in its initial assumptions and in its final results. The data assumed to be *provided* to the learner includes both the motion model and the perceptual model of the robot. These consist of transition and observation probabilities within the grid. Both of these components are *learnt* by our algorithm, although not in a grid context but in a coarser-grained, topological framework. The end result of their algorithm is a probabilistic grid-based map, while ours is a probabilistic topological model, as further explained in the next section.

In addition to being concerned only with locations, rather than with the richer notion of *state*, a fundamental drawback of geometrical maps is their fine granularity and high accuracy. Geometrical maps, particularly grid-based ones, tend to give an accurate and detailed picture of the environment. In cases where it is necessary for a robot to know its exact location in terms of metric coordinates, metric maps are indeed the best choice. However, many planning tasks do not require such fine granularity or accurate measurements, and are better facilitated through a more abstract representation of the world. For example, if a robot needs to deliver a bagel from office *a* to office *b*, all it needs to have is a map depicting the relative location of *a* with respect to *b*, the passageways between the two offices, and perhaps a few other landmarks to help it orient itself if it gets lost. If it has a reasonably well-operating low-level obstacle avoidance mechanism to help it bypass flower pots and chairs that it might encounter on its way, such objects do not need to be part of the environment map. Just as a driver traveling between cities needs to know neither his longitude and latitude coordinates on the globe, nor the location of the specific houses along the way, the robot does not need to know its exact location within the building nor the exact location of various items in the environment, in order to get from one point to another. Hence, the effort of obtaining such detailed maps is not usually justified. In addition the maps can be very large, which makes planning—even though planning is polynomial in the size of the map—inefficient.

## 2.3 Topological Maps and Models

An alternative to the detailed geometric maps are the more abstract topological maps. Such maps specify the *topology* of important landmarks and situations (*states*), and routes or transitions (*arcs*) between them. They are concerned less with the physical location of landmarks, and more with topological relationships between situations. Typically, they are less complex and support much more efficient planning than metric maps. Topological maps are built on lower-level abstractions that allow the robot to move along arcs (perhaps by wall- or road-following), to recognize properties of locations, and to distinguish significant locations as *states*; they are flexible in allowing a more general notion of state, possibly including information about the non-geometrical aspects of the robot's situation.

There are two typical strategies for deriving topological maps: one is to learn the topological map directly; the other is to first learn a geometric map, then to derive a topological model from it through some process of analysis.

A nice example of the second approach is provided by Thrun and Bücken (1996a, 1996b; Thrun, 1999), who use occupancy-grid techniques to build the initial map. This strategy is appropriate when the primary cues for decomposition and abstraction of the map are geometric. However, in many cases, the nodes of a topological map are defined in terms of other sensory data (e.g., labels on a door or whether or not the robot is holding a bagel). Learning a geometric map first also relies on the odometric abilities of a robot; if they are weak and the space is large, it is very difficult to derive a consistent map.

In contrast, our work concentrates on learning a topological model *directly*, assuming that abstraction of the robot's perception and action abilities has already been done. Such abstractions were manually encoded into the lower level of our robot navigational software, as described in Section 6. Work by Pierce and Kuipers (1997) discusses an automatic method for extracting abstract states and features from raw perceptual information.

Kuipers and Byun (1991) provide a strategy for learning *deterministic* topological maps. It works well in domains in which most of the noise in the robot's perception and action is abstracted away, learning from single visits to nodes and traversals of arcs. A strong underlying assumption for these strategies, when building the map, is that the current state can be reliably identified based on local information, or based on distance traversed from the previous well-identified state. These methods are unable to handle situations in which long sequences of actions and observations are necessary to disambiguate the robot's state.

Mataric (1990) provides an alternative approach for learning *deterministic* topological maps, represented as distributed graphs. The learning process again relies on the assumption that the current state can be distinguished from all other states based on local information which includes compass and sonar readings. Uncertainty is not modeled through probability distributions. Instead, matching of current readings to already existing states is not required to be exact, and thresholds of tolerated error are set empirically. Another difference from the work presented here, is that while we learn the complete probabilistic topology of the environment, in Mataric's work the overall topology of the graph is assumed in advance to be a linear list, and additional edges are added during the learning process. No probability distribution is associated with the edges, and a mechanism for choosing which edge to take is determined as part of the goal seeking process, and is not part of the model itself.

Engelson and McDermott (1992) learn "diktiometric" maps (topological maps with metric relations between nodes) from experience. The uncertainty model they use is interval-based rather than probabilistic, and the learned representation is deterministic. *Ad hoc* routines handle problems resulting from failures of the uncertainty representation.

We prefer to learn a combined *model* of the world and the robot's interaction with the world; this allows robust planning that takes into account likelihood of error in sensing and action. The work most closely related to ours is by Koenig and Simmons (1996b, 1996a), who learn POMDP models (stochastic topological models) of a robot hallway environment. They also recognize the difficulty of learning a good model without initial information; they solve the problem by using a human-provided topological map, together with further constraints on the structure of the model. A modified version of the Baum-Welch algorithm learns the parameters of the model. They also developed an incremental version of Baum-Welch that can be used on-line. Their models contain very weak metric information, representing hallways as chains of one-meter segments and allowing the learning algorithm to select the most probable chain length. This method is effective, but results in large models with size proportional to the hallways' length, and strongly depends on the quality of the human-provided initial model.

## 2.4 Learning Automata from Data

Informally speaking, an automaton consists of a set of states and a set of transitions that lead from one state to another. In the context of this work, the automaton states correspond to the states of the modeled environments, and the transitions, to the state changes due to actions performed in the environment. Each transition of the automaton is tagged by a symbol from an input alphabet, $\Sigma$, corresponding to the action or the *input* to the system that caused the state transition. Classical automata theory (e.g., Hopcroft & Ullman, 1979) distinguishes between *deterministic* and *non-deterministic* automata. If, for each alphabet symbol $\alpha$, there is a single edge tagged by it, going out of each state, the automaton is *deterministic*. Otherwise, the transition between states is not uniquely determined by the input symbol and the automaton is *non-deterministic*. If we augment each transition edge of a non-deterministic automaton with a probability of taking it given a certain input, $\alpha$, the resulting automaton is called *probabilistic*.

The basic problem of learning finite *deterministic* automata from given data can be roughly described as follows: *Given a set of positive and a set of negative example strings, $S$ and $T$ respectively, over alphabet $\Sigma$, and a fixed number of states $k$, construct a minimal deterministic finite automaton with no more than $k$ states that accepts $S$ and does not accept $T$.* This problem has been shown to be NP-*complete* (Gold, 1978). Despite the hardness, positive results have been shown possible under various special settings. Angluin (1987) showed that if an oracle can answer membership queries and provide counterexamples to conjectures about the automaton, there is a polynomial time learning algorithm from positive and negative examples. Rivest and Schapire (1987, 1989), provide several effective methods, that under various settings, learn deterministic automata that are correct with *high probability*. While the above work deals with learning from *noise-free* data, Basye, Dean and Kaelbling (1995) presented several algorithms that, with *high probability*, learn input-output deterministic automata, when the data observed by the learner is corrupted by various forms of noise.

In all these cases, the learned automaton is *deterministic* rather than *probabilistic*. The basic learning problem in the probabilistic context is to find an automaton that assigns the same *distribution* as the true one to data sequences, using training data $S$, that was generated by the true automaton. Another form of a learning problem is that of finding a probabilistic automaton $\lambda$ that assigns the maximum likelihood to the training data $S$; that is, an automaton that maximizes $\Pr(S|\lambda)$.

Abe and Warmuth (1992) show that finding a probabilistic automaton with 2 states, even when a small error with respect to the true model is allowed with some probability (the *probably approximately correct*, or *PAC*, learning model), cannot be done in polynomial time with polynomial number of examples, unless NP = RP. From their work arises the broadly accepted conjecture, which has not yet been proven, that learning hidden Markov Models is hard even in the PAC sense. There are two ways to address this hardness: one is to restrict the class of probabilistic models learned, while the other is to learn unrestricted hidden Markov models with good practical results but with no PAC guarantees on the quality of the result.

Work by Ron et al. (1994, 1995, 1998) pursues the first approach, learning restricted classes of automata, namely, *acyclic* probabilistic finite automata, and *probabilistic finite suffix* automata. Both classes are useful for various applications related to natural language processing, and can be learned in polynomial time within the PAC framework.

The second approach, which is the one predominantly taken in this work, is to learn a model that is a member of the complete unrestricted class of hidden Markov models. Only weak guarantees exist about the goodness of the model, but the learning procedure may be directed to obtain practically good results. This approach is based on guessing an automaton (model), and using an iterative procedure to make the automaton fit better to the training data. One algorithm commonly used for this purpose is the Baum-Welch algorithm (Baum, Petrie, Soules, & Weiss, 1970), which is presented in detail by Rabiner (1989). The iterative updates of the model are based on gathering sufficient statistics from the data given the current automaton, and the update procedure is guaranteed to converge to a model that locally maximizes the likelihood function $Pr(\text{data}|\text{model})$. Since the maximum is *local*, the model might not be close enough to the true automaton by which the data was generated, and a challenging problem is to find ways to force the algorithm into converging to higher-likelihood maxima, or at least to make it converge faster, facilitating multiple guesses of initial models, thus raising the probability of converging to higher-likelihood maxima. Such an approach is the one taken in the work presented here.

We assume, throughout this paper, that the *number of states* in the model we are learning is known. This is not a very strong assumption since there are methods for learning the number of states. Regularization methods for deciding on the number of states and other model parameters, are discussed, for instance, in Vapnik's book (1995). We do not address this issue here.

The rest of the work describes our approach to learning topological models. We use noisy odometric information that is readily available in most robots. This geometrical information is typically not used by topological mapping methods. We demonstrate how a topological model and the algorithm used to learn it can be extended to directly incorporate this weak odometric information. We further show that by doing so, we can avoid the use of human-provided *a priori* models and still learn stochastic environment models efficiently and effectively.

# 3. Models and Assumptions

This section describes the formal framework for our work. It starts by introducing the classic hidden Markov model. The model is then extended to accommodate noisy odometric information in its most naïve form, ignoring information about the robot's heading and orientation, and later adapted to accommodate heading information.

We concentrate here on describing models and algorithms for learning HMMs, rather than POMDPs. This means that the robot has no decisions to make regarding its next action at every state; only one action can be executed at each state. In our experiments, a human operator gave the action command associated with each state to the robot when gathering the data. Note that the action is not necessarily the same one for every state, e.g., the robot is told to always turn right in state 1 and move forward at state 2. However, at each state only one action can be taken. The extension to complete POMDPs, which we have implemented, is through learning an HMM for each of the possible actions; it is straightforward although notationally more cumbersome, thus we limit the discussion here to HMMs.

## 3.1 HMMs – The Basics

A hidden Markov model consists of states, transitions, observations and probabilistic behavior, and is formally defined as a tuple $\lambda = \langle S, O, A, B, \pi \rangle$, satisfying the following conditions:

- $S = \{s_0, \ldots, s_{N-1}\}$ is a finite set of $N$ states.

- $O = \{o_0, \ldots, o_{M-1}\}$ is a finite set of $M$ possible observation values.

- $A$ is a stochastic transition matrix, with $A_{i,j} = Pr(q_{t+1} = s_j | q_t = s_i)$, where $0 \leq i, j \leq N-1$. $q_t$ is the state at time $t$. For every state $s_i$, $\sum_{j=0}^{N-1} A_{i,j} = 1$.

  $A_{i,j}$ holds the transition probability from state $s_i$ to state $s_j$.

- $B$ is a stochastic observation matrix, with $B_{j,k} = Pr(v_t = o_k | q_t = s_j)$, where $0 \leq j \leq N-1$, $0 \leq k \leq M-1$. $v_t$ is the observation recorded at time $t$. For every state $s_j$, $\sum_{k=0}^{M-1} B_{j,k} = 1$.

  $B_{j,k}$ holds the probability of observing $o_k$ while being at state $s_j$.

- $\pi$ is a stochastic initial distribution vector, with $\pi_i = Pr(q_0 = s_i)$, $0 \leq i \leq N-1$. $\sum_{i=0}^{N-1} \pi_i = 1$.

  $\pi_i$ holds the probability of being in state $s_i$ at time 0, when starting to record observations.

This model corresponds to a world whose actual state at any given time $t$, $q_t \in S$, is *hidden* and not directly observable, but some observable aspects of the state, $v_t \in O$, are detected and recorded when the state is visited at time $t$. An agent moves from one hidden state to the next according to the probability distribution encoded in matrix $A$. The observed information in each state is governed by the probability matrix $B$. Although our work is concerned with discrete observations, the extension to continuous observations is straightforward and has been well addressed in work on hidden Markov models (Liporace, 1982; Juang, 1985).

Simply stated, the problem of *learning* an HMM is that of "reverse engineering" a hidden Markov model for a stochastic system from the sampled data, generated by the system. We formalize the learning task in Section 4.1. The next section extends HMMs to account for geometric information.

## 3.2 Adding Odometry to Hidden Markov Models

The world is composed of a finite set of states. There is a fundamental distinction in our framework between the term *state* and the term *location*. The *state* of the robot does not directly correspond to its location. A state may include other information, such as the robot's battery level or its *orientation* in that location. A robot standing in the entrance to office 101 facing *right* is in a *different* state than a robot standing in the same place facing *left*; similarly, a robot standing with a bagel in its arm is in a different state from the same robot being in the same position without the bagel.

The dynamics of the world are described by state-transition distributions that specify the probability of making transitions from one state to the next as a result of a certain action. There is a finite set of observations that can be perceived in each state; the relative frequency of each observation is described by a probability distribution and depends only on the current state. In our model, observations are *multi-dimensional*; an observation is a vector of values, each chosen from a finite domain. That is, we *factorize* the observation associated with each state into several components. For instance, as demonstrated in Section 6.1, we view the observation recorded by the robot when standing in an office environment as consisting of three components, corresponding to the three cardinal directions: front, left and right. In this example, the observation vector is thus 3-dimensional. It is assumed that the vector's components are conditionally independent, given the state.

In addition to the above components, each state is assumed to be associated with a *position in a metric space*. Whenever a state transition is made, the robot records an *odometry vector*, which estimates the position of the current state relative to the previous one. For the time being we assume that the odometry vector consists of readings along the $x$ and $y$ coordinates of a global coordinate system, and that these readings are corrupted with independent normal noise. The latter independence assumption is not a strict one, and can be relaxed by introducing a complete covariance matrix, although we have not done this in this work. In Section 3.3 we extend the odometry vector to include information about the *heading* of the robot, and drop the global coordinate framework.

Note that the odometric relationship characterizes a *transition* rather than a state and, as described below, receives a different treatment than the *observations* that are associated with *states*.

There are two important assumptions underlying our treatment of odometric relations between states: First, that there is an inherent "true" odometric relation between the position of every two states in the world; second, that when the robot moves from one state to the next, there is a normal, 0-mean noise around the correct expected odometric reading along each odometric dimension. This noise reflects two kinds of odometric error sources:

- The lack of precision in the discretization of the real world into states (e.g. there is a rather large area in which the robot can stand which can be regarded as "the doorway of the AI lab").

- The lack of precision of the odometric measures recorded by the robot, due to slippage, friction, disalignment of the wheels, imprecision of the measuring instruments, etc.

To formally introduce odometric information into the hidden Markov model framework, we define an *augmented hidden Markov model* as a tuple $\lambda = \langle S, O, A, B, R, \pi \rangle$, where:

- $S = \{s_0, \ldots, s_{N-1}\}$ is a finite set of $N$ states.

- $O = \prod_{i=1}^{l} O_i$ is a finite set of *observation vectors* of length $l$. The $i$th element of an observation vector is chosen from the finite set $O_i$.

- $A$ is a stochastic transition matrix, with $A_{i,j} = Pr(q_{t+1} = s_j | q_t = s_i)$, $0 \leq i, j \leq N-1$.
  $q_t$ is the state at time $t$. For every state $s_i$, $\sum_{j=0}^{N-1} A_{i,j} = 1$.

  $A_{i,j}$ holds the transition probability from state $s_i$ to state $s_j$.

- $B$ is an array of $l$ stochastic observation matrices, with $B_{i,j,k} = Pr(V_t[i] = o_k | q_t = s_j)$; $1 \leq i \leq l$, $0 \leq j \leq N-1$, $o_k \in O_i$; $V_t$ is the observation vector at time $t$; $V_t[i]$ is its $i^{th}$ component.

  $B_{i,j,k}$ holds the probability of observing $o_k$ along the $i^{th}$ component of the observation vector, while being at state $s_j$.

- $R$ is a relation matrix, specifying for each pair of states, $s_i$ and $s_j$, the mean and variance of the $D$-dimensional[2] odometric relation between them. $\mu(R_{i,j}[m])$ is the mean of the $m^{th}$

---

2. For the time being we consider $D$ to be 2, corresponding to $(x, y)$ readings.

component of the relation between $s_i$ and $s_j$ and $\sigma^2(R_{i,j}[m])$, the variance. Furthermore, $R$ is *geometrically consistent*: for each component $m$, the relation $\mu^m(a, b) \stackrel{\text{def}}{=} \mu(R_{a,b}[m])$ must be a *directed metric*, satisfying the following properties for all states $a$, $b$, and $c$:

◇ $\mu^m(a, a) = 0$;

◇ $\mu^m(a, b) = -\mu^m(b, a)$ *(anti-symmetry)*; and

◇ $\mu^m(a, c) = \mu^m(a, b) + \mu^m(b, c)$ *(additivity)* .

This representation of odometric relations reflects the two assumptions, previously stated, regarding the nature of the odometric information. The "true" odometric relation between the position of every two states is represented as the *mean*. The noise around the correct expected odometric relation, accounting for both the lack of precision in the real-world discretization and the inaccuracy in measurement, is represented through the *variance*.

- $\pi$ is a stochastic initial probability vector describing the distribution of the initial state. For simplicity it is assumed here to be of the form $\langle 0, \ldots, 0, 1, 0, \ldots, 0 \rangle$, implying that there is one designated initial state, $s_i$, in which the robot is always started.

This model extends the standard hidden Markov model described in Section 3.1 in two ways:

- It facilitates observations that are factored into components, and represented as vectors. These components are assumed to be conditionally independent of each other given the state. Such factorization, together with the conditional independence assumption, allows for a simple calculation of the probability of the complete observation vector from the probabilities of its components. It therefore results in fewer probabilistic parameters in the learnt model than if we were to view each observation vector, consisting of a possible combination of component-values as a single "atomic" observation.

- It introduces the *odometric relation* matrix $R$ and constraints over its components. Using $R$ and the constraints over it, as explained in Section 4, has proven useful for learning the other model parameters, as demonstrated in Section 6.

### 3.3 Handling Directional Data

We further extend the model to accommodate *directional changes* in addition to the positional changes. There are two issues stemming from directional changes while moving in an environment: the need for non-traditional distributions to model directional changes, and the need to correct for the *cumulative rotational error* which severely interferes with location estimation within a global coordinate framework. A detailed discussion of these two problems and their solution is given in an earlier paper by the authors (Shatkay & Kaelbling, 1998). For the sake of completeness, we briefly review these two issues here.

#### 3.3.1 Circular Distributions

The robot's change in direction as it moves through the environment is expressed in terms of the angular change with respect to its original heading. Since angular measures are inherently circular, treating them as "normally distributed", and using the standard procedures for obtaining sufficient statistics from the data is not adequate. As a trivial example, if we were to average the two angular readings, $173°$ and $-179°$, using simple average we obtain the angle $-3°$, which is far from the intuitive $\sim 180°$, as illustrated in Figure 1.

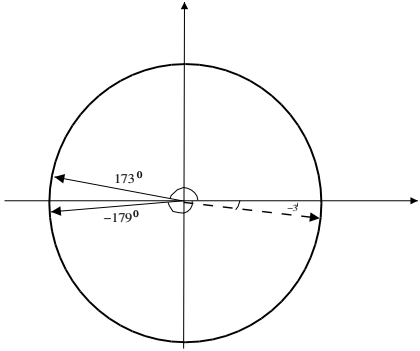

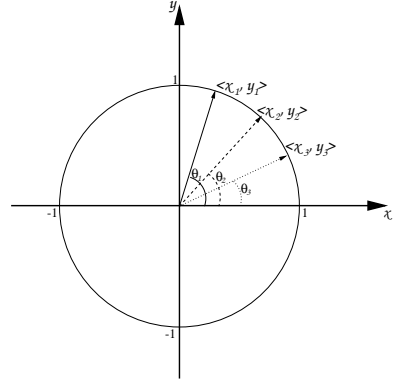

Figure 1: Simple average of two angles, depicted as vectors to the unit circle. The average angle is formed by the dashed vector.

Figure 2: Directional data represented as angles and as vectors on the unit circle.

To address the circularity issue, we use the *von Mises* distribution, which is a circular version of the normal distribution, to model the change in heading between two states, as explained below.

A collection of changes in heading within a two dimensional space can be represented in terms of either Cartesian or polar coordinates. Using a Cartesian system, $n$ changes in headings can be recorded as a sequence of 2-dimensional vectors, $(\langle x_1, y_1 \rangle, \ldots \langle x_n, y_n \rangle)$, on the unit circle, as shown in Figure 2. The same changes can also be represented as the corresponding angles between the radii from the center of the unit circle and the $X$ axis, $(\theta_1, \ldots, \theta_n)$, respectively. The relationship between the two representations is:

$$x_i = \cos(\theta_i), \qquad y_i = \sin(\theta_i), \quad (1 \le i \le n) .$$

The vector mean of the $n$ points, $\langle \overline{x}, \overline{y} \rangle$, is calculated as:

$$\overline{x} = \frac{\sum_{i=1}^{n} \cos(\theta_i)}{n}, \qquad \overline{y} = \frac{\sum_{i=1}^{n} \sin(\theta_i)}{n} . \tag{1}$$

Using polar coordinates, we can express the mean vector in terms of angle, $\overline{\theta}$, and length, $\overline{a}$, where (except for the case $\overline{x} = \overline{y} = 0$):

$$\overline{\theta} = \arctan(\frac{\overline{y}}{\overline{x}}), \qquad \overline{a} = (\overline{x}^2 + \overline{y}^2)^{\frac{1}{2}} .$$

The angle $\overline{\theta}$ is the mean angle, while the length $\overline{a}$ is a measure (between 0 and 1) of how concentrated the sample angles are around $\overline{\theta}$. The closer $\overline{a}$ is to 1, the more concentrated the sample is around the mean, which corresponds to a smaller sample variance.

Intuitively, a satisfactory circular version of the normal distribution would have a mean for which the maximum likelihood estimate is the average angle as calculated above. In a way analogous to Gauss' derivation of the Normal distribution, von Mises developed such a circular version (Gumbel, Greenwood, & Durand, 1953; Mardia, 1972), which is defined as follows:

**Definition:** *A circular random variable, $\theta$, $0 \le \theta \le 2\pi$, is said to have the von Mises distribution with parameters $\mu$ and $\kappa$, where $0 \le \mu \le 2\pi$ and $\kappa > 0$, if its probability density*

*function is:*

$$f_{\mu,\kappa}(\theta) = \frac{1}{2\pi I_0(\kappa)} e^{\kappa \cos(\theta - \mu)} \, ,$$

*where $I_0(\kappa)$ is the modified Bessel function of the first kind and order 0:*

$$I_0(\kappa) = \sum_{r=0}^{\infty} \frac{1}{r!^2} (\frac{1}{2}\kappa)^{2r} \, . \tag{2}$$

The parameters $\mu$ and $\kappa$ correspond to the distribution's *mean* and *concentration* respectively.

While other circular-normal distributions do exist, the von Mises has the desirable estimation procedure alluded to earlier: Given a set of heading samples, angles $\theta_1, \ldots \theta_n$, from a von Mises distribution, the maximum likelihood estimate $\overline{\mu}$ for $\mu$ is:

$$\overline{\mu} = \arctan(\frac{\overline{y}}{\overline{x}}) \, ,$$

where $\overline{y}$, $\overline{x}$ are as defined in Equation 1.

The maximum likelihood estimate for the concentration parameter, $\kappa$, is the $\overline{\kappa}$ that satisfies:

$$\frac{I_1(\overline{\kappa})}{I_0(\overline{\kappa})} = \max[\frac{1}{n}\sum_{i=1}^{n} \cos(\theta_i - \mu), \ 0] \, ,$$

where $I_1$ is the modified Bessel function of the first kind and order 1:

$$I_1(\kappa) = \sum_{r=0}^{\infty} \frac{1}{r!(r+1)!} (\frac{1}{2}\kappa)^{2r+1} \, . \tag{3}$$

Further information about the estimation procedure is beyond the scope of this paper and can be found elsewhere (Gumbel et al., 1953; Mardia, 1972).

To conclude, we assume that the change in heading $\Delta\theta$ is *von Mises*-distributed, around a mean $\mu$ with concentration parameter $\kappa$. This assumption is reflected in the model learning procedures as explained later in Section 4.2.3. The change in heading $\langle \mu^\theta(a,b), \kappa^\theta(a,b) \rangle$ between each pair of states $(a,b)$ completes the set of parameters included in the relation matrix $R$ which was introduced earlier in Section 3.2.

### 3.3.2 Cumulative Rotational Error

We tend to think about an environment as consisting of landmarks fixed in a global coordinate system and corridors or transitions connecting these landmarks. This idea underlies the typical maps constructed and used in everyday life. However, this view of the environment may be problematic when robots are involved.

Conceptually, a robot has two levels at which it operates; the *abstract* level, in which it centers itself through corridors, follows walls and avoids obstacles, and the *physical* level in which motors turn the wheels as the robot moves. In the physical level many inaccuracies can manifest themselves: wheels can be unaligned with each other resulting in a drift to the right or to the left, one motor can be slightly faster than another resulting in similar drifts, an obstacle under one of the wheels can cause the robot to rotate around itself slightly, or uneven floors may cause the robot to slip in a certain direction. In addition, the measuring instrumentation for odometric information may not be accurate in and of itself. At the abstract level, corrective actions are constantly executed to overcome the physical drift and drag. For example, if the left wheel is misaligned and drags the robot leftwards, a corrective action of moving to the right is constantly taken in the higher level to keep the robot centered in the corridor.

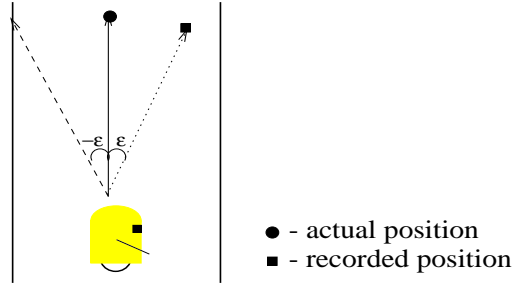

Figure 3: A robot moving along the solid arrow, while correcting for drift in the direction of the dashed arrow. The dotted arrow marks its *recorded* change in position.

The phenomena described above have a significant effect on the odometry recorded by the robot, if such data interpreted with respect to one global framework. For example, consider the robot depicted in Figure 3. It drifts to the left $-\epsilon°$ when moving from one state to the next, and corrects for it by moving $\epsilon°$ to the right in order to maintain itself centered in the corridor.

Let us assume that states are 5 meters apart along the center of the corridor, and that the center of the corridor is aligned with the $Y$ axis of the global coordinate system. The robot steps back and forth in the corridor from one state to the next. Whenever the robot reaches a state, its odometry reading changes by $\langle x, y, \theta \rangle$ along the $\langle X, Y, \text{heading} \rangle$ dimensions, respectively. As the robot proceeds, the deviation with respect to the $X$ axis becomes more and more severe. Thus, after going through several transitions, the odometric changes recorded between every pair of states, if taken with respect to a global coordinate system, become larger and larger. Similar problems of inconsistent odometric changes recorded between pairs of states can arise along any of the odometric dimensions. It is especially severe when such inconsistencies arise with respect to the heading, since this can lead to mistakenly switching movement along the $X$ and the $Y$ axes, as well as confusion between forwards and backwards movement (when the deviation in the heading is around $90°$ or $180°$ respectively).

In early work (Shatkay & Kaelbling, 1997) we assumed *perpendicularity* of the corridors, which was taken advantage of while the robot collected the data. Odometric readings were recorded with respect to a *global coordinate system*, and the robot could re-align itself with the origin after each turn. A trajectory of odometry recorded under this *perpendicularity assumption* by our robot Ramona, along the $x$ and $y$ axes is given in Figure 4. The sequence shown was recorded while the robot drove repeatedly around a loop of corridors. Further details about the data gathering process are provided in Section 6. In contrast, Figure 5 shows a trajectory of another sequence of odometric readings recorded by Ramona, driving through the same corridors, *without* using the perpendicularity assumption. The data collected under the latter setting is subjected to *cumulative rotational error*.

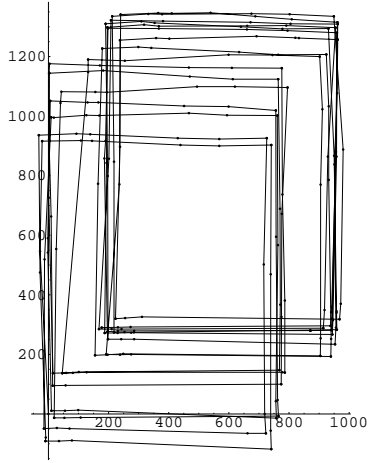

Figure 4: Sequence gathered by Ramona, perpendicularity assumed.

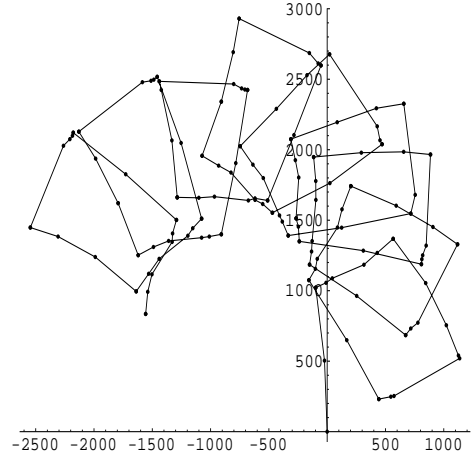

Figure 5: Sequence gathered by Ramona, no perpendicularity assumed.

Such data can be handled through *state-relative coordinate systems* (Shatkay & Kaelbling, 1998). The latter implies that each state $s_i$ has its own coordinate system, as shown in Figure 6: the origin is anchored in $s_i$, the $Y$ axis is aligned with the robot's heading in the state (denoted by bold arrows in the figure), and the $X$ axis is perpendicular to it. This is in contrast to a global coordinate system which is anchored in the initial starting state. Within the global coordinate system, the relations recorded may vary greatly among multiple instances of the same transition between the same pair of states. By using the state-relative system, the recorded and learned relationship between each pair of states, $\langle s_i, s_j \rangle$, is reliable, despite the fact that it is based on multiple transitions recorded from $s_i$ to $s_j$.

Under state-relative coordinate systems, the geometric relation stored in $R_{ij}$, (which was introduced in Section 3.2), is expressed for each pair of states, $s_i$ and $s_j$, with respect to the coordinate system associated with state $s_i$. Accordingly, the constraints imposed over the $x$ and $y$ components of the relation matrix must be specified with respect to the explicit coordinate system used, as explained below.

Given a pair of states $a$ and $b$, we denote by $\mu^{\langle x,y \rangle}(a, b)$ the vector $\langle \mu(R_{a,b}[x]), \mu(R_{a,b}[y]) \rangle$. Let us define $\mathcal{T}_{ab}$ to be the transformation that maps an $\langle x_a, y_a \rangle$ point represented with respect to the coordinate system of state $a$, to the same point represented with respect to the coordinate system of state $b$, $\langle x_b, y_b \rangle$.

More explicitly, let $\mu^\theta_{ab}$ be the mean change in heading from state $a$ to state $b$. Applying $\mathcal{T}_{ab}$ to a vector $\langle \begin{smallmatrix} x_a \\ y_a \end{smallmatrix} \rangle$ results in the vector $\langle \begin{smallmatrix} x_b \\ y_b \end{smallmatrix} \rangle$ as follows:

$$\left\langle \begin{matrix} x_b \\ y_b \end{matrix} \right\rangle = \mathcal{T}_{ab} \left\langle \begin{matrix} x_a \\ y_a \end{matrix} \right\rangle = \left\langle \begin{matrix} x_a \cos(\mu^\theta_{ab}) - y_a \sin(\mu^\theta_{ab}) \\ x_a \sin(\mu^\theta_{ab}) + y_a \cos(\mu^\theta_{ab}) \end{matrix} \right\rangle \quad .$$

The consistency constraints within this framework must be restated as:

⋄ $\mu^{\langle x,y \rangle}(a, a) = \langle 0, 0 \rangle$;

⋄ $\mu^{\langle x,y \rangle}(a, b) = -\mathcal{T}_{ba}[\mu^{\langle x,y \rangle}(b, a)]$ *(anti-symmetry)*;

⋄ $\mu^{\langle x,y \rangle}(a, c) = \mu^{\langle x,y \rangle}(a, b) + \mathcal{T}_{ba}[\mu^{\langle x,y \rangle}(b, c)]$ *(additivity)*.

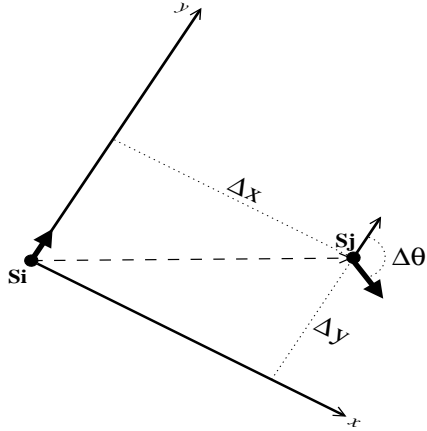

Figure 6: A robot in state $S_i$, faces in the $Y$-axis direction; the relation $S_i, S_j$ is WRT $S_i$'s coordinate system.

These consistency constraints are the ones that need to be enforced by our learning algorithm which constructs the HMM. It is important to note that the transformation $\mathcal{T}$ itself does not constitute a set of additional parameters that need to be learnt. Rather, it is calculated in terms of the heading-change parameter, $\mu^\theta$, which is already an integral part of the relation matrix we have defined in Sections 3.2 and 3.3.1.

We have introduced the basic formal model that we use for representing environments and the robot's interaction with them. In the following section we state the learning problem and describe the basic algorithm for learning the model from data.

## 4 Learning HMMs with Odometric Information

This section formalizes the learning problem for HMMs, and discusses how odometric information is incorporated into the learning algorithm. An overview of the complete algorithm is provided in the Appendix for this paper.

### 4.1 The Learning Problem

The learning problem for hidden Markov models can be generally stated as follows: *Given an experience sequence* E, *find a hidden Markov model that could have generated this sequence and is "useful" or "close to the original" according to some criterion.* An explicit common statistical approach is to look for a model $\lambda$ that maximizes the *likelihood* of the data sequence E given the model. Formally stated, it maximizes $\Pr(E|\lambda)$. However, given the complicated landscape of typical likelihood functions in a multi-parameter domain, obtaining a maximum likelihood model is not feasible. All studied practical methods, and in particular the well-known Baum-Welch algorithm (Rabiner (1989) and references therein) can only guarantee a *local-maximum* likelihood model.

Another way of evaluating the quality of a learned model is by comparing it to the true model. We note that stochastic models (such as HMMs) induce a probability distribution over all observation sequences of a given length. The Kullback-Leibler (Kullback & Leibler, 1951) divergence of a learned distribution from a true one is a commonly used measure for estimating how good a learned model is. Obtaining a model that minimizes this measure is a possible learning goal. The culprit here is that in practice, when we learn a model from data, we do not have any "ground truth" model to compare the learned model with. Still, we can evaluate *learning algorithms* by measuring how well they perform on data obtained from known models. It is reasonable to expect that an algorithm that learns well from data that is generated from a model we do have, will perform well on data generated from an unknown model, assuming that the models indeed form a suitable representation of the true generating process. We discuss the Kullback-Leibler (KL) divergence in more detail in Section 6.2 in the context of evaluating our experimental results.

To summarize, the learning problem as we address it in this work is that of obtaining a model by attempting to (locally) maximize the likelihood, while evaluating the results based on the KL-divergence with respect to the true underlying distribution, when such a distribution is available.

## 4.2 The Learning Algorithm

The learning algorithm starts from an initial model $\lambda_0$ and is given an *experience* sequence E; it returns a revised model $\lambda$, which (locally) maximizes the likelihood $P(\mathsf{E}|\lambda)$. The experience sequence E is of length $T$; each element, $\mathsf{E}_t$, for $0 \leq t \leq (T-1)$, is a pair $\langle r_t, V_t \rangle$, where $r_t$ is the observed relation vector along the $x$, $y$ and $\theta$ dimensions, between the states $q_{t-1}$ and $q_t$, and $V_t$ is the observation vector at time $t$.

Our algorithm extends the standard Baum-Welch algorithm to deal with the relational information and the factored observation sets. The Baum-Welch algorithm is an expectation-maximization (EM) algorithm (Dempster, Laird, & Rubin, 1977); it alternates between

- the *E-step* of computing the state-occupation and state-transition probabilities, $\gamma$ and $\xi$, at each time in the sequence given E and the current model $\lambda$, and
- the *M-step* of finding a new model, $\overline{\lambda}$, that maximizes $P(\mathsf{E}|\lambda, \gamma, \xi)$,

providing monotone convergence of the likelihood function $P(\mathsf{E}|\lambda)$ to a local maximum.

However, our extension introduces an additional component, namely, the relation matrix $R$. It can be viewed as having two kinds of observations: *state* observations (as the ordinary HMM — with the distinction that we observe integer vectors rather than integers) and *transition* observations (the odometry relations between states). The latter must satisfy geometrical constraints. Hence, an extension of the standard update formulae, as described below, is required.

### 4.2.1 State-Occupation Probabilities

Following Rabiner (1989), we first compute the forward ($\alpha$) and backward ($\beta$) matrices. $\alpha_t(i)$ denotes the probability density value of observing $\mathsf{E}_0$ through $\mathsf{E}_t$ and $q_t = s_i$, given $\lambda$; $\beta_t(i)$ is the probability density of observing $\mathsf{E}_{t+1}$ through $\mathsf{E}_{T-1}$ given $q_t = s_i$ and $\lambda$. Formally:

$$\alpha_t(i) = Pr(\mathsf{E}_0, \ldots, \mathsf{E}_t, q_t = s_i | \lambda) \ ,$$

$$\beta_t(i) = Pr(\mathsf{E}_{t+1}, \ldots, \mathsf{E}_{T-1} | q_t = s_i, \lambda) \ .$$

When some of the measurements are continuous (as is the case with $R$), these matrices contain probability density values rather than probabilities.
The forward procedure for calculating the $\alpha$ matrix is initialized with

$$\alpha_0(i) = \begin{cases} b_0^i & \text{if } \pi_i = 1 \\ 0 & \text{otherwise} \ , \end{cases}$$

and continued for $0 < t \leq T-1$ with

$$\alpha_t(j) = \sum_{i=0}^{N-1} \alpha_{t-1}(i) A_{i,j} f(r_t | R_{i,j}) b_t^j \quad . \tag{4}$$

The expression $f(r_t | R_{i,j})$ denotes the *density* at point $r_t$ according to the distribution represented by the means and variances in entry $i,j$ of the relation matrix $R$, while $b_t^j$ is the probability of observing vector $v_t$ in state $s_j$; that is, $b_t^j = \prod_{i=0}^{l} B_{i,j,v_t[i]}$.

The backward procedure for calculating the $\beta$ matrix is initialized with $\beta_{T-1}(j) = 1$, and continued for $0 \leq t < T-1$ with

$$\beta_t(i) = \sum_{j=0}^{N-1} \beta_{t+1}(j) A_{i,j} f(r_{t+1} | R_{i,j}) b_{t+1}^j \quad . \tag{5}$$

Given $\alpha$ and $\beta$, we now compute for each given time point $t$ the state-occupation and state-transition probabilities, $\gamma$ and $\xi$. The state-occupation probabilities, $\gamma_t(i)$, representing the probability of being in state $s_i$ at time $t$ given the experience sequence and the current model, are computed as follows:

$$\gamma_t(i) = \Pr(q_t = s_i | \mathsf{E}, \lambda) = \frac{\alpha_t(i)\beta_t(i)}{\sum_{j=0}^{N-1} \alpha_t(j)\beta_t(j)} \quad . \tag{6}$$

Similarly, $\xi_t(i,j)$, the state-transition probabilities from state $i$ to state $j$ at time $t$ given the experience sequence and the current model, are computed as:

$$\begin{aligned}
\xi_t(i,j) &= \Pr(q_t = s_i, q_{t+1} = s_j | \mathsf{E}, \lambda) \\
&= \frac{\alpha_t(i) A_{i,j} b_{t+1}^j f(r_{t+1} | R_{i,j}) \beta_{t+1}(j)}{\sum_{i=0}^{N-1} \sum_{j=0}^{N-1} \alpha_t(i) A_{i,j} b_{t+1}^j f(r_{t+1} | R_{i,j}) \beta_{t+1}(j)} \quad .
\end{aligned} \tag{7}$$

These are essentially the same formulae appearing in Rabiner's tutorial (Rabiner, 1989), but they also take into account the density of the odometric relations.

In the next phase of the algorithm, the goal is to find a new model, $\overline{\lambda}$, that maximizes the likelihood conditioned on the current transition and observation probabilities, $\Pr(\mathsf{E} | \lambda, \gamma, \xi)$. Usually, this is simply done using maximum-likelihood estimation of the probability distributions in $A$ and $B$ by computing expected transition and observation frequencies. In our model we must also compute a new relation matrix, $R$, under the constraint that it remain geometrically consistent. Through the rest of this section we use the notation $\overline{v}$ to denote a reestimated value, where $v$ denotes the current value.

### 4.2.2 Updating Transition and Observation Parameters

The $A$ and $B$ matrices can be straightforwardly reestimated. $\overline{A}_{i,j}$ is the expected number of transitions from $s_i$ to $s_j$ divided by the expected number of transitions from $s_i$, and $\overline{B}_{i,j,k}$ is the expected number of times $o_k$ is observed along the $i$th dimension when in state $s_j$, divided by the expected number of times of being in $s_j$:

$$\overline{A}_{i,j} = \frac{\sum_{t=0}^{T-2} \xi_t(i,j)}{\sum_{t=0}^{T-2} \gamma_t(i)} \quad , \quad \overline{B}_{i,j,k} = \frac{\sum_{t=0}^{T-1} \delta_{[V_t[i]=o_k]} \gamma_t(j)}{\sum_{t=0}^{T-1} \gamma_t(i)} \quad . \tag{8}$$

The expression $\delta_c$ denotes an indicator function with value 1 if condition $c$ is true and 0 otherwise.

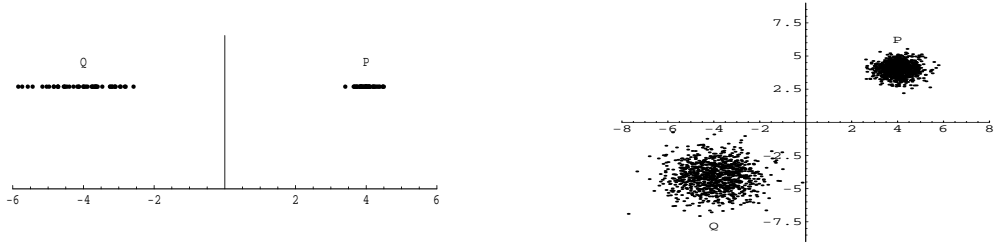

Figure 7: Examples of two sets of normally distributed points with constrained means, in 1 and in 2 dimensions.

### 4.2.3 UPDATING RELATION PARAMETERS

When reestimating the relation matrix, $R$, the geometrical constraints induce interdependencies among the optimal mean estimates as well as between optimal variance estimates and mean estimates. Parameter estimation under this form of constraints is almost untreated in mainstream statistics (Bartels, 1984) and we found no previous existing solutions to the estimation problem addressed here. As an illustration for the issues involved in estimation under constraints consider the following estimation problem of 2 normal means:

**Example 4.1** *The data consists of two sample sets of points* $P = \{p_1, p_2, \ldots, p_n\}$ *and* $Q = \{q_1, q_2, \ldots, q_k\}$, *independently drawn from two distinct normal distributions with means* $\mu_P$, $\mu_Q$ *and variances* $\sigma_P^2$, $\sigma_Q^2$, *respectively. We are asked to find maximum likelihood estimates for the two distribution parameters. Moreover, we are told that the means of the two distributions are related, such that* $\mu_Q = -\mu_P$, *as illustrated in Figure 7. If not for the latter constraint, the task is simple (DeGroot, 1986), and we have:*

$$\mu_P = \frac{\sum_{i=1}^n p_i}{n} , \ \sigma_P^2 = \frac{\sum_{i=1}^n (p_i - \mu_P)^2}{n} ,$$

*and similarly for* $\mu_Q$ *and* $\sigma_Q^2$. *However, the constraint* $\mu_P = -\mu_Q$ *requires finding a single mean,* $\mu$, *and setting the other one to its negated value,* $-\mu$. *Intuitively, when choosing such a maximum likelihood single mean, the more concentrated sample should have more effect, while the more varied sample should be more "submissive." Thus, the overall sample deviation from the means would be minimized and the likelihood of the data maximized. Therefore, there is a mutual dependence between the estimation of the mean and the estimation of the variance.*

*Since the samples are independently drawn, their joint likelihood function is:*

$$f(P, Q | \mu_P, \mu_Q, \sigma_P^2, \sigma_Q^2) = \prod_{i=1}^n \frac{e^{\frac{-(p_i - \mu_P)^2}{2\sigma_P^2}}}{\sqrt{2\pi}\sigma_P} \cdot \prod_{j=1}^k \frac{e^{\frac{-(q_j - \mu_Q)^2}{2\sigma_Q^2}}}{\sqrt{2\pi}\sigma_Q} .$$

*By taking the derivatives of this joint log-likelihood function, with respect to* $\mu_P$, $\sigma_P$ *and* $\sigma_Q$, *and equating them to 0, while using the constraint* $\mu_Q = -\mu_P$, *we obtain the following set of mutual equations for maximum likelihood estimators:*

$$\mu_P = \frac{(\sigma_Q^2 \sum_{i=1}^n p_i) - (\sigma_P^2 \sum_{j=1}^k q_j)}{n\sigma_Q^2 + k\sigma_P^2}, \ \ \mu_Q = -\mu_P,$$

$$\sigma_P^2 = \frac{\sum_{i=1}^n (p_i - \mu_P)^2}{n}, \ \ \sigma_Q^2 = \frac{\sum_{j=1}^k (q_j + \mu_P)^2}{k} .$$

*By substituting the expressions for $\sigma_P$ and $\sigma_Q$ into the expression for $\mu_P$, we obtain a cubic equation which is cumbersome, but still solvable (in this simple case). The solution provides a maximum likelihood estimate for the mean and variance under the constraint $\mu_Q = -\mu_P$.* □

We now proceed to the actual update of the relation matrix under constraints. For clarity, we initially discuss only the first two geometrical constraints, and discuss the additivity constraint in Section 4.3. Recall that we concentrate here on the enforcement of *global constraints,* appropriate under the *perpendicularity assumption,* although the same idea is applied in the case of state-relative constraints.

*Zero distances* between states and themselves are trivially enforced, by setting all the diagonal entries in the $R$ matrix to 0, with a small variance.

*Anti-symmetry* within a global coordinate system is enforced by using the data recorded along the transition from state $s_j$ to $s_i$ as well as from state $s_i$ to $s_j$ when reestimating $\mu(R_{i,j})$. As demonstrated in Example 4.1, the variance has to be taken into account, leading to the following set of mutual equations:

$$\overline{\mu}_{i,j}^m \;=\; \frac{\sum_{t=0}^{T-2}\left[\frac{r_t[m]\xi_t(i,j)}{(\overline{\sigma}_{i,j}^m)^2} - \frac{r_t[m]\xi_t(j,i)}{(\overline{\sigma}_{j,i}^m)^2}\right]}{\sum_{t=0}^{T-2}\left[\frac{\xi_t(i,j)}{(\overline{\sigma}_{i,j}^m)^2} + \frac{\xi_t(j,i)}{(\overline{\sigma}_{j,i}^m)^2}\right]} \;,\tag{9}$$

$$(\overline{\sigma}_{i,j}^m)^2 \;=\; \frac{\sum_{t=0}^{T-2}[\xi_t(i,j)(r_t[m] - \overline{\mu}_{i,j}^m)^2]}{\sum_{t=0}^{T-2}\xi_t(i,j)} \;.\tag{10}$$

For the $x$ and $y$ dimensions, $(m = x, y)$, this amounts to a complicated but still solvable cubic equation. However, in the more general case, when accounting for the orientation of the robot, and also when complete additivity is enforced, we do not obtain such closed form reestimation formulae.

To avoid these hardships, we use a *lag-behind* update rule; the yet-unupdated estimate of the variance is used for calculating a new estimate for the mean, and this new mean estimate is used to update the variance, using Equation 10.[3] Thus, the mean is updated using a variance parameter that *lags behind* it in the update process, and the reestimation Equation (9) needs to use $\sigma^m$ rather than $\overline{\sigma}^m$ as follows:

$$\overline{\mu}_{i,j}^m = \frac{\sum_{t=0}^{T-2}\left[\frac{r_t[m]\xi_t(i,j)}{(\sigma_{i,j}^m)^2} - \frac{r_t[m]\xi_t(j,i)}{(\sigma_{j,i}^m)^2}\right]}{\sum_{t=0}^{T-2}\left[\frac{\xi_t(i,j)}{(\sigma_{i,j}^m)^2} + \frac{\xi_t(j,i)}{(\sigma_{j,i}^m)^2}\right]} \;.\tag{11}$$

As we have shown (Shatkay, 1999), this lag-behind policy is an instance of generalized EM (McLachlan & Krishnan, 1997). The latter guarantees monotone convergence to a local maximum of the likelihood function, even when each "maximization" step *increases* rather than strictly maximizes the expected likelihood of the data given the current model.

Similarly, the reestimation formula for the von Mises *mean* ($\mu$) and *concentration* ($\kappa$) parameters of the *heading change* between states $s_i$ and $s_j$ is the solution to the equations:

$$\overline{\mu}_{i,j}^\theta \;=\; \arctan\left(\frac{\sum_{t=0}^{T-2}[\sin(r_t[\theta])(\xi_t(i,j)\overline{\kappa}_{i,j} - \xi_t(j,i)\overline{\kappa}_{j,i})]}{\sum_{t=0}^{T-2}[\cos(r_t[\theta])(\xi_t(i,j)\overline{\kappa}_{i,j} + \xi_t(j,i)\overline{\kappa}_{j,i})]}\right)$$

---

3. A similar approach, termed *one step late* update, is taken by others applying EM to highly non-linear optimization problems (McLachlan & Krishnan, 1997).

$$\frac{I_1[\overline{\kappa}_{i,j}^\theta]}{I_0[\overline{\kappa}_{i,j}^\theta]} = \max\left[\frac{\sum_{t=0}^{T-2}[\xi_t(i,j)\cos(r_t[\theta] - \overline{\mu}_{i,j}^\theta)]}{\sum_{t=0}^{T-2}\xi_t(i,j)},\ 0\right]\ , \tag{12}$$

where $I_0$ and $I_1$ are the modified Bessel functions as defined by Equations 2 and 3 in Section 3.3.1.

Again, to avoid the need to solve the mutual equations, we take advantage of the lag-behind strategy, updating the mean using the current estimates of the concentration parameters, $\kappa_{i,j}$, $\kappa_{j,i}$, as follows:

$$\overline{\mu}_{i,j}^\theta = \arctan\left(\frac{\sum_{t=0}^{T-2}[\sin(r_t[\theta])(\xi_t(i,j)\kappa_{i,j} - \xi_t(j,i)\kappa_{j,i})]}{\sum_{t=0}^{T-2}[\cos(r_t[\theta])(\xi_t(i,j)\kappa_{i,j} + \xi_t(j,i)\kappa_{j,i})]}\right)\ , \tag{13}$$

and then calculating the new concentration parameters based on the newly updated mean, as the solution to Equation 12, through the use of lookup-tables.

A possible alternative to our lag-behind approach is to update the mean as though the assumption $\sigma_{j,i} = \sigma_{i,j}$ holds. Under this assumption, the variance terms in Equation 9 cancel out, and the mean update is independent of the variance once again. Then the variances are updated as stated in Equation 10, *without* assuming any constraints over them. This approach was taken in earlier stages of this work (Shatkay & Kaelbling, 1997, 1998). The lag-behind strategy is superior, both according to our experiments, and due to its being an instance of generalized EM.

### 4.3 Enforcing Additivity

Note that the additivity constraint directly implies the other two geometrical constraints[4]. Thus, enforcing it results in complete geometrical consistency. We present here the method for directly enforcing additivity through the reestimation procedure along the $x$ and $y$ dimensions. For the heading dimension we describe how complete geometrical consistency is achieved through the projection of anti-symmetric estimates onto a geometrically-consistent space. As before, to simplify the presentation, we focus on the case of global coordinate systems. The same basic idea applies to state-relative coordinate systems, but the relationship used to recover the mean $\mu_{ij}$ from individual state coordinates is more complex.

#### 4.3.1 Additivity in the $x$, $y$ dimensions

The main observation underlying our approach is that the additivity constraint is a result of the fact that states can be embedded in a *geometrical space*. That is, assuming we have $N$ states, $s_0, \ldots, s_{N-1}$, there are points on the $X$, $Y$ and $\theta$ axes, $x_0, \ldots, x_{N-1}, y_0, \ldots, y_{N-1}, \theta_0, \ldots, \theta_{N-1}$, respectively, such that each state, $s_i$, is associated with the coordinates $\langle x_i, y_i, \theta_i \rangle$. Assuming one global coordinate system, the mean odometric relation from state $s_i$ to state $s_j$ can be expressed as: $\langle x_j - x_i, y_j - y_i, \theta_j - \theta_i \rangle$.

During the *maximization* phase of the EM iteration, rather than try to maximize with respect to $N^2$ *odometric relation vectors*, $\langle \mu_{ij}^X, \mu_{ij}^Y, \mu_{ij}^\theta \rangle$, we *reparameterize* the problem. Specifically, we express each odometric relation as a function of two of the $N$ *state positions*, and maximize with respect to the *unconstrained, $N$ state positions*. For instance, for the $X$ dimension, rather than search for $N^2$ maximum likelihood estimates for $\mu_{ij}^x$, we use the maximization step to find $N$ 1-dimensional points, $x_0, \ldots, x_{N-1}$. We can then calculate $\mu_{ij}^x = x_j - x_i$. Moreover, since all we are interested in is finding the best *relationships* between $x_i$ and $x_j$, we can fix one of the $x_i$'s at 0 (e.g. $x_0 = 0$), and find optimal estimates for the remaining $N-1$ state positions. The variance reestimation remains as before, and the lag-behind policy is used to eliminate the interdependency between the update of the mean and the variance parameters.

---

4. $\{\mu(a,a) = \mu(a,a) + \mu(a,a)\} \Rightarrow (\mu(a,a) = 0)$ ; $\{(\mu(a,a) = 0)\,;\,(\mu(a,a) = \mu(a,b) + \mu(b,a))\} \Rightarrow (\mu(a,b) = -\mu(b,a))$.

### 4.3.2 Additive Heading Estimation

Unfortunately, the reparameterization described above is not feasible for estimation of changes in *heading*, due to the von Mises distribution assumption over the heading measures. By reparameterizing $\mu_{ij}^\theta$ as $\theta_j - \theta_i$ and trying to maximize the likelihood function with respect to the $\theta$ parameters, we obtain a set of $N-1$ *trigonometric equations* with terms of the form $\cos(\theta_j) \cdot \sin(\theta_i)$ which do not enable simple solution.

As an alternative, it is possible to use the anti-symmetric reestimation procedure described earlier, followed by a perpendicular *projection* operator, mapping the resulting headings vector $\langle \mu_{00}^\theta, \ldots, \mu_{ij}^\theta, \ldots, \mu_{N-1,N-1}^\theta \rangle$, $0 \leq i, j \leq N-1$, which does not satisfy additivity, onto a vector of headings within an *additive linear vector space*. Simple orthogonal projection is not satisfactory within our setting, since it simply looks for the additive vector closest to the non-additive one. This procedure ignores the fact that some of the entries in the non-additive vector are based on a lot of observations, and are therefore more reliable, while other, less reliable ones, are based on hardly any data at all. Intuitively, we would like to keep the estimates that are well accounted for intact, and adapt the less reliable estimates to meet the additivity constraint. More precisely, there are heading-change estimates between states that are better accounted for than others, in the sense that the transitions between these states have higher expected counts than transition between other states (higher $\sum_t \xi_t(i,j)$). We would like to project the *non-additive* heading estimates vector onto a *subspace* of the *additive* vector space, in which the vectors have the same values as the non-additive vector in the entries that are well-accounted for, that is, those with the highest values of $\sum_t \xi_t(i,j)$. The difficulty is that the latter subspace is *not a linear* vector space (for instance, it does not satisfy closure under scalar multiplication), and the projection operator over linear spaces cannot be applied directly. Still, this set of vectors does form an *affine* vector space, and we can project onto it using an algebraic technique, as explained below.[5]

**Definition** $\mathsf{A} \subseteq \mathcal{R}^n$ is an *n-dimensional* affine *space if for all vectors* $v_a \in \mathsf{A}$, *the set of vectors:* $\mathsf{A} - v_a \overset{def}{=} \{u_a - v_a | u_a \in \mathsf{A}\}$ *is a* linear *space.*

Hence, we can pick a vector in an affine space, $v_{a_1} \in \mathsf{A}$, and define the translation $T_a : \mathsf{A} \to V$, where $V$ is a linear space, $V = \mathsf{A} - v_{a_1}$. This translation is trivially extended for any vector $v' \in \mathcal{R}^n$, by defining $T_a(v') = v' - v_{a_1}$. In order to project any vector $v \in R^n$ onto $\mathsf{A}$, we apply the translation $T_a$ to $v$ and project $T_a(v)$ onto $V$, which results in a vector $\mathcal{P}(T_a(v))$ in $V$. By applying the inverse transform $T_a^{-1}$ to it, we obtain the projection of $v$ on $\mathsf{A}$, as demonstrated in Figure 8. The linear space in the figure is the two dimensional vector space $\{\langle x, y \rangle | y = -x\}$, and the affine space is $\{\langle x, y \rangle | y = -x + 4\}$. The transform $T_a$ consists of subtracting the vector $\langle 0, 4 \rangle$. The solid arrow corresponds to the direct projection of the vector $v$ onto the point $\mathcal{P}(v)$ of the affine space. The dotted arrows represent the projection via translation of $v$ to $T_a(v)$, the projection of the latter onto the linear vector space, and the inverse translation of the result, $\mathcal{P}(T_a(v))$, onto the affine space.

---

5. Many thanks to John Hughes for introducing us to this technique.

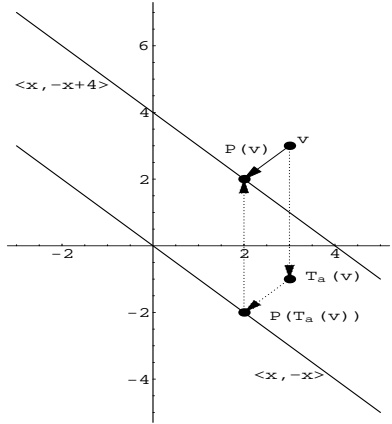

Figure 8: Projecting $v$ onto the affine vector space $\{\langle x, y\rangle \mid y = -x + 4\}$.

Although the procedure for preserving additivity over headings is not formally proven to preserve monotone convergence of the likelihood function towards a local maximum, our extensive experiments consisting of hundreds of runs have shown that monotone convergence is preserved.

## 5 Choosing an Initial Model

Typically, in instances of the Baum-Welch algorithm, an initial model is picked uniformly at random from the space of all possible models, perhaps trying multiple initial models to find different local likelihood maxima. An alternative approach we have reported (Shatkay & Kaelbling, 1997) was based on clustering the accumulated odometric information using the simple k-means algorithm (Duda & Hart, 1973), taking the clusters to be the states in which the observations were recorded, to obtain state and observation counts and estimate the model parameters.

If perpendicularity is assumed when collecting the data, as shown in Figure 4, the k-means algorithm assigns the same cluster (state) to odometric readings recorded at close locations, leading to reasonable initial models. However, when this assumption is dropped, as illustrated in Figure 5, the cumulative rotational error distorts the odometric location recorded within a global coordinate system, so that the location assigned to the same state during multiple visits varies greatly and would not be recognized as "the same" by a simple location-based clustering algorithm. To overcome this, we developed an alternative initialization heuristics, which we call *tag-based initialization*. It is based *directly* on the recorded *relations* between states, rather than on states' absolute location. For clarity, the description here consists mostly of an illustrative example, and concentrates on the case where global consistency constraints are enforced.

Given a sequence of observations and odometric readings E, we begin by clustering the odometric readings into *buckets*. The number of buckets is at most the number of distinct state transitions recorded in the sequence. The goal at this stage is to have each bucket contain all the odometric readings that are close to each other along all three dimensions.

To achieve this, we start by fixing a predetermined, small standard deviation value along the $x$, $y$, and $\theta$ dimensions. Denote these standard deviation values $\sigma_x, \sigma_y, \sigma_\theta$ respectively, (typically $\sigma_x = \sigma_y$). The first odometric reading is assigned to bucket 0 and the mean of this bucket is set to be the value of this reading. Through the rest of the process the subsequent odometric readings are examined. If the next reading is within 1.5 standard deviations along each of the three dimensions from the mean of some existing non-empty bucket, add it to the bucket and update the bucket mean accordingly. If not, assign it to an empty bucket and set the mean of the bucket to be this reading.

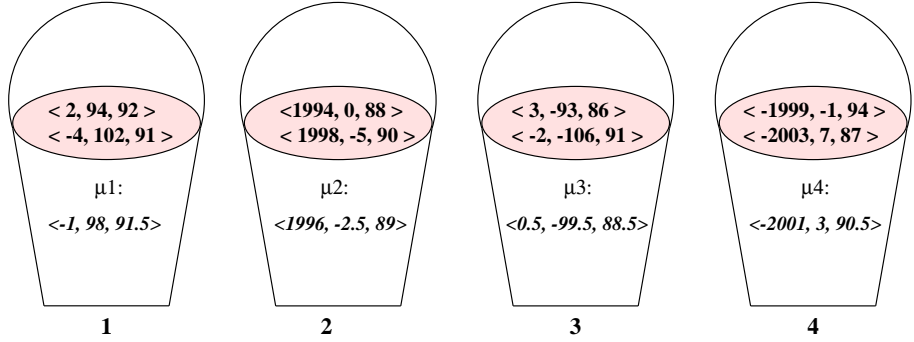

Figure 9: The bucket assignment of the example sequence.

Intuitively, by using this heuristic each of the resulting buckets is tightly concentrated about its mean. We note that other clustering algorithms (Duda & Hart, 1973) could be used at the bucketing stage.

**Example 5.1** *We would like to learn a 4-state model from a sequence of odometric readings, $\langle x, y, \theta \rangle$ as follows:*

$$\langle 2\ 94\ 92 \rangle, \ \langle 1994\ 0\ 88 \rangle, \ \langle 3\ -93\ 86 \rangle, \ \langle -1999\ 1\ 94 \rangle,$$
$$\langle -4\ 102\ 91 \rangle, \ \langle 1998\ -5\ 90 \rangle, \ \langle -2\ -106\ 91 \rangle, \ \langle -2003\ 7\ 87 \rangle \,.$$

*As a first stage we place these readings into buckets. Suppose the standard deviation constant is 20. The placement is as shown in Figure 9. The mean value associated with each bucket is shown as well.* □

The next stage of the algorithm is the *state-tagging* phase, in which each odometric reading, $r_t$, is assigned a pair of states, $s_i, s_j$, denoting the origin state (from which the transition took place) and the destination state (to which the transition led), respectively. In conjunction, the mean entries, $\mu_{ij}$, of the relation matrix, $R$, are populated.

**Example 5.1 (cont.)** *Returning to the sequence above, the process is demonstrated in Figure 10. We assume that the data recording starts at state 0, and that the odometric change through self transitions is 0, with some small standard deviation (we use 20 here as well). This is shown on part A of the figure.*

*Since the first element in the sequence, $\langle 2\ 94\ 92 \rangle$, is more than two standard deviations away from the mean $\mu[0][0]$ and no other entry in the relation row of state 0 is populated, we pick 1 as the next state and populate the mean $\mu[0][1]$ to be the same as the mean of bucket 1, to which $\langle 2\ 94\ 92 \rangle$ belongs. To maintain geometrical consistency the mean $\mu[1][0]$ is set to $-\mu[0][1]$, as shown in part B of the figure. We now have populated 2 off-diagonal entries, and the state sequence is $\langle 0, 1 \rangle$. The entry $[0][1]$ in the matrix becomes associated with bucket 1, and this information is recorded for helping with tagging future odometric readings belonging to the same bucket.*

*The next odometric reading, $\langle 1994\ 0\ 88 \rangle$, is a few standard deviations from any populated mean in row 1 (where 1 is the current believed state). Hence, we pick a new state 2, and set the mean $\mu[1][2]$ to be $\mu 2$—the mean of bucket 2—to which the reading belongs (Figure 10 C). The entry $[1][2]$ is recorded as associated with bucket 2. To preserve anti-symmetry and additivity, $\mu[2][1]$ is set to $-\mu[1][2]$. $\mu[0][2]$ is set to be the sum $\mu[0][1] + \mu[1][2]$, and $\mu[2][0]$ is set to $-\mu[0][2]$.*

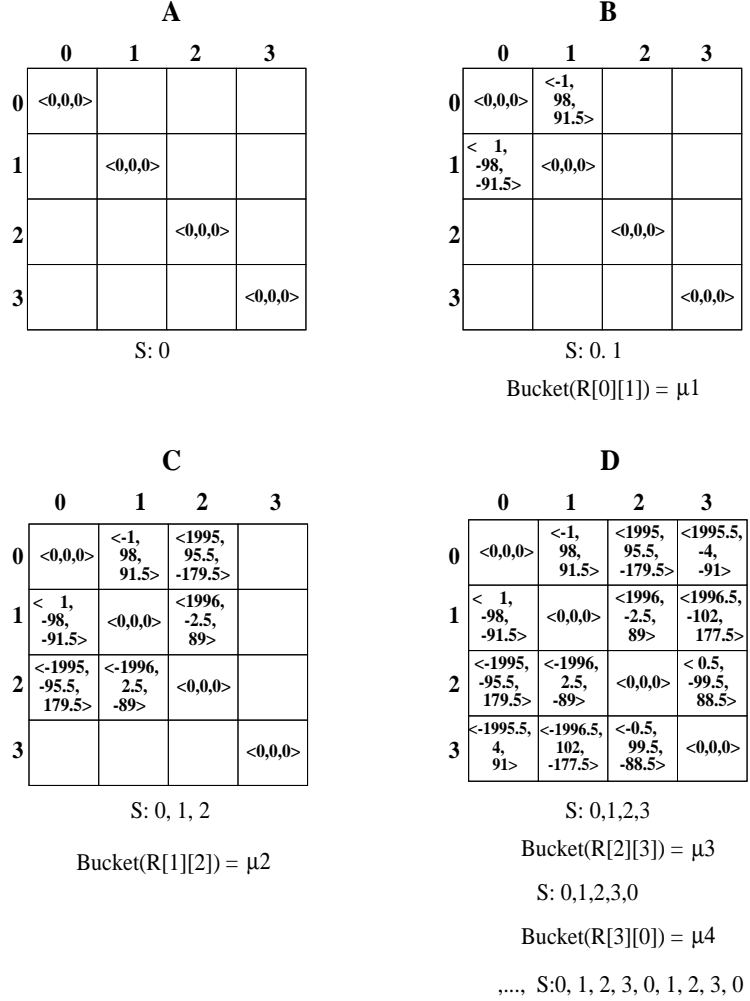

Figure 10: Populating the odometric relation matrix and creating a state tagging sequence.

*Similarly, $\mu[2][3]$ is updated to be the mean of bucket 3, causing the setting of $\mu[3][2]$, $\mu[1][3]$, $\mu[0][3]$, $\mu[3][1]$, and $\mu[3][0]$. Bucket 3 is associated with $\mu[2][3]$.*

*At this stage the odometric table is fully populated, as shown in part D of Figure 10. The state sequence at this point is: $\langle 0, 1, 2, 3 \rangle$. The next reading, $\langle -1999 \ -1 \ 94 \rangle$, is within one standard deviation from $\mu[3][0]$ and therefore the next state is 0. Entry $[3][0]$ is associated with bucket 4, (the bucket to which the reading was assigned), and the state sequence becomes: $\langle 0, 1, 2, 3, 0 \rangle$.*

*The next reading, being from bucket 1, is associated with the relation from state 0 that is tagged by bucket 1, namely, state 1. By repeating this for the last two readings, the final state transition sequence becomes $\langle 0, 1, 2, 3, 0, 1, 2, 3, 0 \rangle$.* □

Note that the process described in the above illustration was simplified. In the general case, we need to take into account the rotational error in the data, use state-relative coordinate systems, and therefore populate the entries under the transformed anti-symmetry and additivity constraints:

⋄ $\mu^{\langle x,y \rangle}(a, b) = -\mathcal{T}_{ba}[\mu^{\langle x,y \rangle}(b, a)]$ ;
⋄ $\mu^{\langle x,y \rangle}(a, c) = \mu^{\langle x,y \rangle}(a, b) + \mathcal{T}_{ba}[\mu^{\langle x,y \rangle}(b, c)]$,

as defined in Section 3.3.2.

It is possible that by the end of the tagging algorithm, some rows or columns of the relation matrix are still unpopulated. This happens when there is too little data to learn from or when the number of states provided to the algorithm is too large with respect to the actual model. In such cases we can either "trim" the model, using the number of populated rows as the number of states, or pick random odometric readings to populate the rest of the table, improving these estimates later. Note that the first approach suggests a method for *learning the number of states* in the model when this is not given, starting from a gross over-estimate of the number, and truncating it to the number of populated rows in the odometric table after initialization is performed.

Once the state-transition sequence is obtained, the rest of the initialization algorithm is the same as it is for k-means based initialization, deriving state-transition counts from the state-transition sequence, assigning the observations to the states under the assumption that the state sequence is correct, and obtaining state-transition and observation probabilities. The initialization phase does not incur much computational overhead, and is equivalent time-wise to performing one additional iteration of the EM procedure.

# 6 Experiments and Results

The goal of the work described so far is to use odometry to improve the learning of topological models, while using fewer iterations and less data. We tested our algorithm in a simple robot-navigation world. Our experiments consist of running the algorithm both on data obtained from a simulated model and on data gathered by our mobile robot, Ramona. The amount of data gathered by Ramona is used here as a proof of concept but is not sufficient for statistical analysis. For the latter, we use data obtained from the simulated model. We gathered data and used the algorithms both with and without the perpendicularity assumption (see Section 3.3.2), and results are provided from both settings.

## 6.1 Robot Domain

The robot used in our experiments, Ramona, is a modified RWI B21 robot. It has a cylindrical synchro-drive base, 24 ultrasonic sensors and 24 infrared sensors, situated evenly around its circumference. The infrared sensors are used mostly for short-range obstacle avoidance. The ultrasonic sensors are longer ranged, and are used for obtaining (noisy) observations of the environment. In the experiments described here, the robot follows a *prescribed* path through the corridors in the office environment of our department. Thus, there is no decision-making involved, and an HMM is a sufficient model, rather than a complete POMDP.

Low-level software[6] provides a level of abstraction that allows the robot to move through hallways from intersection to intersection and to turn ninety degrees to the left or right. The software uses sonar data to distinguish *doors, openings,* and *intersections* along the path, and to stop the robot's current action whenever such a landmark is detected. Each stop—either due to the natural termination of an action or due to a landmark detection—is considered by the robot to be a "state".

At each stop, ultrasonic data interpretation allows the robot to perceive, in each of the three cardinal directions, (front, left and right), whether there is an open space, a door, a wall, or something unknown.

Encoders on the robot's wheels allow it to estimate its pose (position and orientation) with respect to its pose at the previous intersection. After recording both the sonar-based observations and the odometric information, the robot goes on to execute the next prescribed action. The action command is issued manually by a human operator. Of course, both the action performance and the perception routines are subject to error. The path Ramona followed consists of 4 connected corridors in our building, which include 17 states, as shown in Figure 11.

---

6. The low-level software was written and maintained by James Kurien.

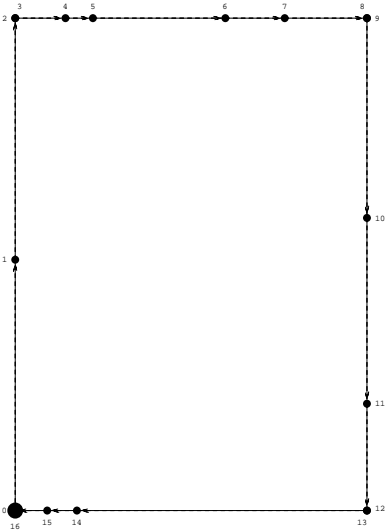

Figure 11: True model of the corridors Ramona traversed. Arrows represent the prescribed path direction.

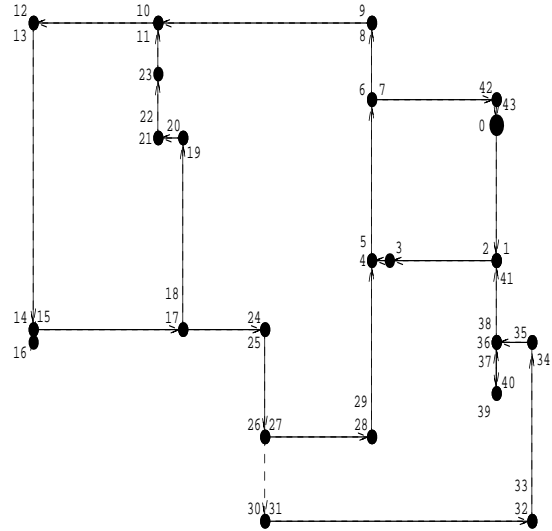

Figure 12: True model of a prescribed path through the simulated hallway environment.

In our simulation, we manually generated an HMM representing a prescribed path of the robot through the complete office environment of our department, consisting of 44 states, and the associated transition, observation, and odometric distributions. The transition probabilities reflect an action failure rate of about $5 - 10\%$. That is, the probability of moving from the current state to the correct next state in the environment, under the predetermined action is between 0.85 and 0.95. The probability of self transition is typically between 0.05 and 0.15. Some small probability (typically smaller than 0.02) is sometimes assigned to other transitions. Our experience with the real robot proves that this is a reasonable transition model, since typically the robot moves to the next state correctly, and the only error that occurs with some significant frequency is when it does not move at all, due to sonar interpretation indicating a barrier when there is actually none. Once the action command is repeated the robot usually performs the action correctly, moving to the expected next state. The observation distribution typically assigns probabilities of $0.85 - 0.95$ to the true observation that should be perceived by the robot at each state, and probabilities of $0.05 - 0.15$ to other observations that might be perceived. For example, if a door should actually be perceived, a door is typically assigned a probability of $0.85 - 0.9$, a wall is assigned a probability of $0.09 - 0.1$ and an open space is assigned a probability of about 0.01 to be perceived. The standard deviation around odometric readings is about 5% of the mean.

Figure 12 shows the HMM corresponding to the simulated hallway environment. Observations and orientation are omitted from the figure for clarity. Nodes correspond to states in the environment, while directed edges correspond to the corridors; the arrows point at the direction in which the corridors were traversed. Further interpretation of the figures is provided in the following section.

## 6.2 Evaluation Method

There are a number of different ways of evaluating the results of a model-learning algorithm. None are completely satisfactory, but they all give some insight into the utility of the results.

In this domain, there are transitions and observations that usually take place, and are therefore more likely than the others. Furthermore, the relational information gives us a rough estimate of the metric locations of the states. To get a *qualitative* sense of the plausibility of a learnt model, we can extract an *essential* map from the learnt model, consisting of the *states*, the most *likely transitions* and the *metric measures* associated with them, and ask whether this map corresponds to the *essential* map underlying the true world.

Figures 11 and 12 are such essential versions of the true models, while Figures 15 and 17, shown later, are essential versions of representative learnt ones (obtained from sequences gathered under the perpendicularity assumption). Black dots represent the physical locations of states, and each state is assigned a unique number. Multiple state numbers associated with a single location typically correspond to different orientations of the robot at that location. The larger black circle represents the *initial* state. Solid arrows represent the most likely non-self transitions between the states. Dashed arrows represent the other transitions when their probability is 0.2 or higher. Typically, due to the predetermined path we have taken, the connectivity of the modeled environment is low, and therefore the transitions represented by dashed arrows are *almost* as likely as the most likely ones. Note that the length of the arrows, within each plot, is significant and represents the length of the corridors, drawn to scale.

It is important to note that the figures do not provide a complete representation of the models. First, they lack observation and orientation information. We stress the fact that the figures serve more as a visual aid than as a plot of the true model. We are looking for a good *topological* model rather than a *geometrical* model. The figures provide a geometrical embedding of the topological model. However, even when the geometry, as described by the relation matrix, is different, the topology, as described by the transition and observation matrices, can still be valid.

Traditionally, in simulation experiments, the learnt model is *quantitatively* compared to the actual model that generated the data. Each of the models induces a probability distribution on strings of observations; the asymmetric Kullback-Leibler divergence (Kullback & Leibler, 1951) between the two distributions is a measure of how good the learnt model is with respect to the true model. Given a true probability distribution $P = \{p_1, ..., p_n\}$ and a learnt one $Q = \{q_1, ..., q_n\}$, the KL divergence of $Q$ with respect to $P$ is:

$$D(P||Q) \stackrel{\text{def}}{=} \sum_{i=1}^{n} p_i \log_2 \frac{p_i}{q_i} \quad .$$

We report our results in terms of a sampled version of the KL divergence, as described by Juang and Rabiner (1985). It is based on generating sequences of sufficient length (5 sequences of 1000 observations in our case) according to the distribution induced by the true model, and comparing their log-likelihood according to the learnt model with the true model log-likelihood. The total difference in log-likelihood is then divided by the total number of observations, accumulated over all the sequences, giving a number that roughly measures the difference in log-likelihood per observation. Formally stated, let $M_1$ be the true model and $M_2$ a learnt one. By generating $K$ sequences $S_1, \ldots, S_K$, each of length $T$, from the true model, $M_1$, the sampled KL-divergence, $D_s$ is:

$$D_s(M_1||M_2) = \frac{\sum_{i=1}^{K} [\log(\Pr(S_i|M_1)) - \log(\Pr(S_i|M_2))]}{KT} \quad .$$

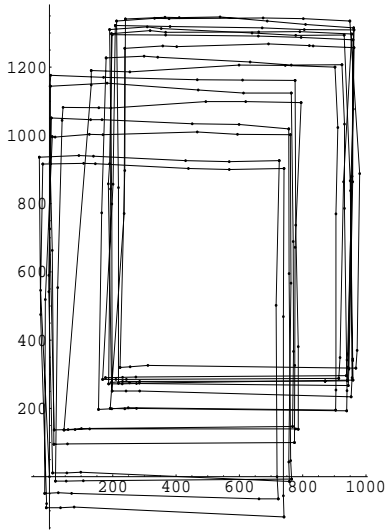

Figure 13: Sequence gathered by Ramona, perpendicularity assumed.

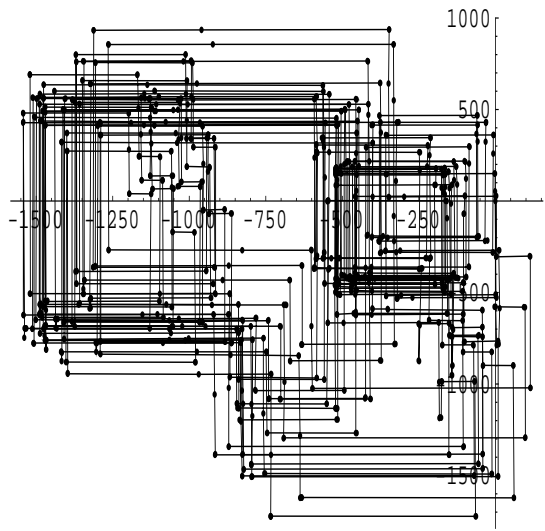

Figure 14: Sequence generated by our simulator, perpendicularity assumed.

We ignore the odometric information when applying the KL measure, thus allowing comparison between purely topological models that are learnt with and without odometry.

## 6.3 Results within a Global Framework

We let Ramona go around the path depicted in Figure 11 and collect a sequence of about 300 observations, while assuming *perpendicularity* of the environment, that is, at every turning point the angle of turn is 90°. Thus at each turn Ramona realigns its odometric readings with its initial $X$ and $Y$ axes. Figure 13 plots the sequence of metric coordinates, gathered in this way, while accumulating consecutive odometric readings, projected on $\langle x, y \rangle$. We applied the learning algorithm to the data 30 times. 10 of these runs were started from a k-means-based initial model, 10 started from a tag-based initial model, and 10 started from a random initial model. In addition we also ran the standard Baum-Welch algorithm, ignoring the odometric information, 10 times. (Note that there is non-determinism even when using biased initial models, since the k-means clustering starts from random seeds, and low[7] random noise is added to the data in all algorithms to avoid numerical instabilities, thus multiple runs give multiple results). We report here the results obtained using the tag-based method, which is the most appropriate initialization method in the general case. These results are contrasted with those obtained when odometric information is not used at all. For a comparison of all four settings the reader is referred to the complete report of this work (Shatkay, 1999).

Figure 15 shows the essential representations of typical learnt models starting from a tag-based initial model. The geometry of the learnt model strongly corresponds to that of the true environment, and most of the states' positions were learnt correctly. Although the figure does not show it, the learnt observation distributions at each state usually match well with the true observations.

To demonstrate the effect of odometry on the quality of the learnt topological model, we contrast the plotted models learnt using odometry with a representative *topological* model learnt *without*

---

7. A random number between -1cm and 1cm is added to recorded distances that are typically several meters long.

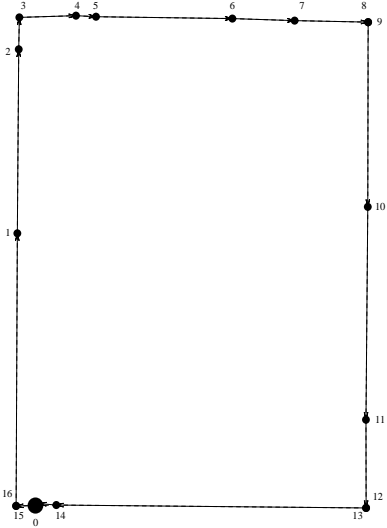

Figure 15: Learnt model of the corridors Ramona traversed.

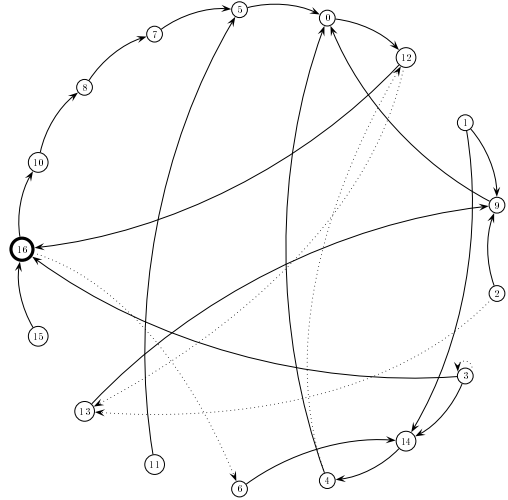

Figure 16: The topology of a model learnt without the use of odometry.

*the use of odometric information.* Figure 16 shows the topology of a typical model learnt without the use of odometric information. In this case, the arcs represent only topological relationships, and their length is not meaningful. The initial state is shown as a bold circle. It is clear that the topology learnt does not match the characteristic loop topology of the true environment.

For obtaining statistically sufficient information, we generated 5 data sequences, each of length 1000, using Monte Carlo sampling from the hidden Markov model whose projection is shown in Figure 12. One of these sequences is depicted in Figure 14. The figure demonstrates that the noise model used in the simulation is indeed compatible with the noise pattern associated with real robot data. We used four different settings of the learning algorithm:

- starting from a biased, tag-based, initial model and using odometric information;

- starting from a biased, k-means-based, initial model and using odometric information;

- starting from an initial model picked uniformly at random, while using odometric information;

- starting from a random initial model *without* using odometric information (standard Baum-Welch).

For each sequence and each of the four algorithmic settings we ran the algorithm 10 times. To keep the discussion focused, we concentrate here on the first and the last of these settings and the reader is referred to a more extensive report (Shatkay, 1999) for a complete discussion.

In all the experiments, $N$ was set to be 44, which is the "correct" number of states; for generalization, it will be necessary to use cross-validation or regularization methods to select model complexity. Section 5 also suggests one possible heuristic for obtaining an estimate of the number of states.

Figure 17 shows an essential version of one learnt model, obtained from the sequence shown in Figure 14, using tag-based initialization. We note that the learnt model is not completely accurate with respect to the true model. However, there is an obvious correspondence between groups of states in the learnt and true models, and most of the transitions (as well as the observations, which are not shown) were learnt correctly. The quality of the geometry of the learnt model in this simulated large environment varies, and the geometrical results are not as uniformly good as was the case when learning the smaller environment from real robot data. As the environment gets large, the global relations between remote states, which are reflected in the geometrical consistency constraints, become harder to learn. Still, the topology of the learnt model as demonstrated by our statistical experiments is good.

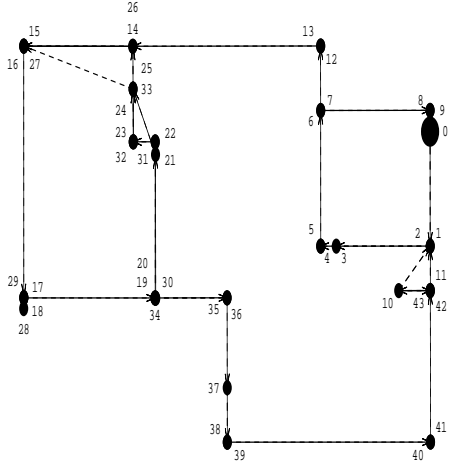

Figure 17: Learnt model of the simulated hallway environment.

Table 1 lists the KL divergence between the true and learnt model, as well as the number of runs until convergence was reached, for each of the 5 sequences for both the setting that uses odometric information under tag-based initialization and the learning algorithm that does not use odometric information, averaged over 10 runs per sequence. We stress that each KL divergence measure is calculated based on *new data sequences* that are generated from the true model, as described in Section 6.2. The 5 sequences from which the models were learnt *do not* participate in the testing process.

The KL divergence with respect to the true model for models learnt using odometry, is about *5-6 times smaller* than for models learnt *without* odometric data. The standard deviation around the means is about 0.2 for KL distances for models learnt with odometry and 1.5 for the no-odometry setting. To check the significance of our results we used the simple two-sample t-test. The models learnt using odometric information have statistically significantly ($p \ll 0.0005$) lower average KL divergence than the others.

| Seq. # | | 1 | 2 | 3 | 4 | 5 |
|---|---|---|---|---|---|---|
| **With** | KL | 0.981 | 1.290 | 1.115 | 1.241 | 1.241 |
| **Odo** | **Iter #** | 16.70 | 20.90 | 22.30 | 12.70 | 27.50 |
| **No** | KL | 6.351 | 4.863 | 5.926 | 6.261 | 4.802 |
| **Odo** | **Iter #** | 124.1 | 126.0 | 113.0 | 107.4 | 122.9 |

Table 1: Average results of two learning settings with five training sequences.

In addition, the number of iterations required for convergence when learning using odometric information is roughly 4-5 times smaller than that required when ignoring such information. Again, the t-test verifies the significance of this result.

Under all three initialization settings, the models learnt are topologically somewhat inferior (and this is with high statistical significance), in terms of the KL divergence, to those learnt without enforcing additivity, reported in earlier papers (Shatkay & Kaelbling, 1997, 1998). This is likely to be a result of the very strong constraints enforced during the learning process, which prevent the algorithm from searching better areas of the learning-space, and restrict it to reach poor local maxima. The geometry looks superior in some cases, but it is not significantly better. However, there seems to be less variability in the quality of the geometrical models across multiple runs when additivity is enforced.

While the details of an extensive comparison between the different initialization methods are beyond the scope of this paper, we point out that our studies of both small and large models show that when large models and long data sequences are involved, random initialization often results in lower KL-divergence than the tag-based initialization. This again has to do with the strong bias of tag-based initialization, which can lead to very peaked models compared with the less-peaked distributions associated with the true model. Random initialization leads to flatter models. As the KL-divergence strongly penalizes models that are much more peaked than the true ones, randomly initialized models are often closer, in terms of this measure, to the true models than the very peaked ones learnt from other initial models. When learning small models, where sufficient training data is available, the tag-based initialization results in models that are clearly superior to the random ones. Again, the reader is referred to the complete report of this work (Shatkay, 1999) for a comparative study of all initialization methods under the various settings.

## 6.4 Results within a Relative Framework

We applied the algorithm described in Section 4.3, extended to accommodate the state-relative constraints (as listed in Section 3.3.2). The data used was gathered by the robot from the same environment, and generated from the same simulated model as before (Figures 11, 12). However, here the data is generated *without assuming perpendicularity*. This means that the $x$ and $y$ coordinates are not realigned after each turn with the global $x$ and $y$ axes, but rather, recorded "as-is." The evaluation methods stay as described above.

Figure 18 shows the projection of the odometric readings that Ramona recorded along the $x$ and $y$ dimensions, while traversing this environment. For obtaining statistically sufficient information, we generated 5 data sequences, each of length 800, using Monte Carlo sampling from the hidden Markov model whose projection is shown in Figure 12. One of these sequences is depicted in Figure 19.

Figure 20 shows a typical model obtained by applying the algorithm enforcing the complete geometrical consistency, to the robot data shown in Figure 18, using tag-based initialization. We note that the rectangular geometry of the environment is preserved, although state 0 does not participate in the loop. This is explained by observing the corresponding area of the true environment as depicted in Figure 11, consisting of the 4 states clustered at the bottom left corner (*0, 14, 15* and *16*). Due to the relatively large number of states that are close together in that area of the true environment, it was not recognized that we ever returned particularly to state 0 during the loop. Therefore, there was only one transition recorded from state 0 to state

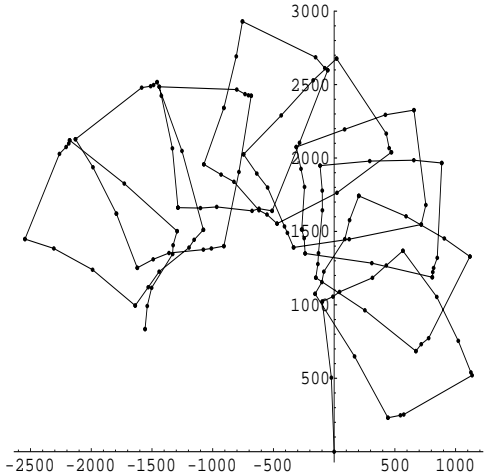

Figure 18: Sequence gathered by Ramona, no perpendicularity assumed.

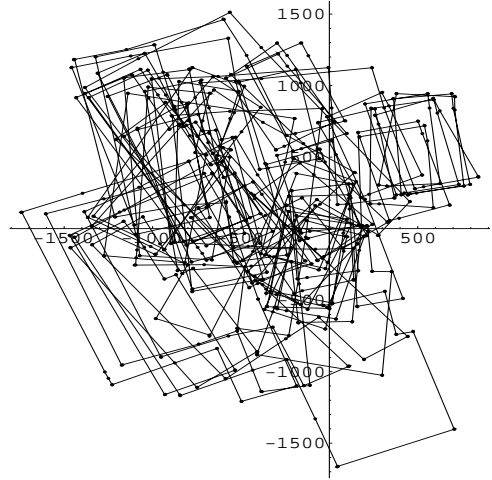

Figure 19: Sequence generated by our simulator, no perpendicularity assumed.

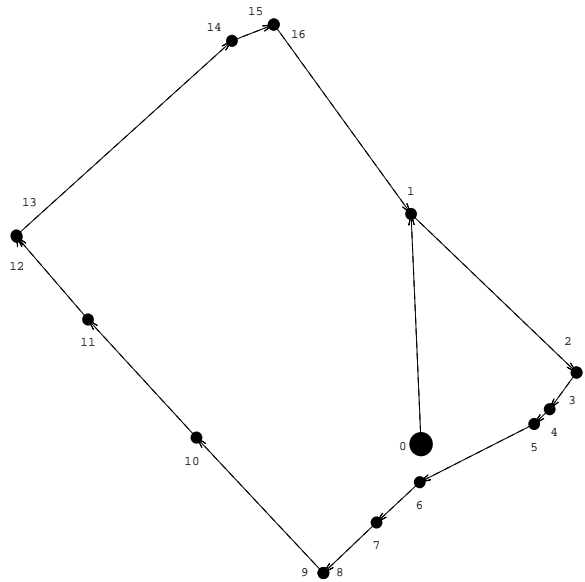

Figure 20: Learnt model of the corridors Ramona traversed. Initialization is tag-based.

1 according to the expected transition counts calculated by the algorithm. When projecting the angles to maintain additivity, (as described in Section 4.3.2), the angle from state 0 to 1 was therefore compromised, allowing geometrical consistency to maintain the rectangular geometry among the more regularly visited states.

For the purpose of quantitatively evaluating the learning algorithm we list in Table 2 the KL divergence between the true and learnt model, as well as the number of iterations until convergence was reached, for each of the 5 simulation sequences with/without odometric information, averaged over 10 runs per sequence. The table demonstrates that the KL divergence with respect to the true model for models learnt using odometric data, is about *8 times smaller* than for models learnt without it. To check the significance of our results we again use the simple two-sample t-test. The models learnt using odometric information have highly statistically significantly ($p \ll 0.0005$) lower average KL divergence than the others. In addition, the number of iterations required for convergence when learning using odometric information is smaller than required when ignoring such information. Again, the t-test verifies the significance ($p < 0.005$) of this result.

| Seq. # | | 1 | 2 | 3 | 4 | 5 |
|---|---|---|---|---|---|---|
| **With** | KL | 1.46 | 1.18 | 1.20 | 1.02 | 1.22 |
| **Odo** | **Iter #** | 11.8 | 36.8 | 30.7 | 24.6 | 33.3 |
| **No** | KL | 6.91 | 9.93 | 10.03 | 9.54 | 12.43 |
| **Odo** | **Iter #** | 113.3 | 113.1 | 102.0 | 104.2 | 112.5 |

Table 2: Average results of 2 learning settings with 5 training sequences.

It is important to point out that the number of iterations, although much lower, does not automatically imply that our algorithm runs in less time than the non-odometric Baum-Welch. The major bottleneck is caused by the need to compute within the forward-backward calculations, as described in Section 4.2.1, the values of the normal and the von-Mises densities. These require the calculation of exponent terms rather than simple multiplications, slowing down each iteration, under the current naïve implementation. However, we can solve this by augmenting the program with look-up tables for obtaining the relevant values rather than calculating them. In addition, we can take advantage of the symmetry in the relations table to cut down on the amount of calculation required. It is also possible to use the fact that many odometric relations remain unchanged (particularly in the later iterations of the algorithm) from one iteration to the next, and therefore values can be cached and shared between iterations rather than be recalculated at each iteration.

## 6.5 Reducing the Amount of Data

Learning HMMs obviously requires visiting states and transitioning between them multiple times, to gather sufficient data for robust statistical estimation. Intuitively, exploiting odometric data can help reduce the number of visits needed for obtaining a reliable model.

To examine the influence of reduction in the length of data sequences on the quality of the learnt models, we took one of the 5 sequences and used its prefixes of length 100 to 800 (the complete sequence), in increments of 100, as training sequences. We ran the two algorithmic settings over each of the 8 prefix sequences, 10 times repeatedly. We then used the KL-divergence as described above to evaluate each of the resulting models with respect to the true model. For each prefix length we averaged the KL-divergence over the 10 runs.

The plot in Figure 21 depicts the average KL-divergence as a function of the sequence length for each of the two settings. It demonstrates that, in terms of the KL divergence, our algorithm, which uses odometric information, is robust in the face of data reduction, (down to 200 data points). In contrast, learning without the use of odometry quickly deteriorates as the amount of data is reduced.

We note that the data sequence is twice as "wide" when odometry is used than when it is not; that is, there is more information in each element of the sequence when odometry data is recorded. However, the effort of recording this additional odometric information is negligible, and is well rewarded by the fact that fewer observations and less exploration are required for obtaining a data sequence sufficient for adequate learning.

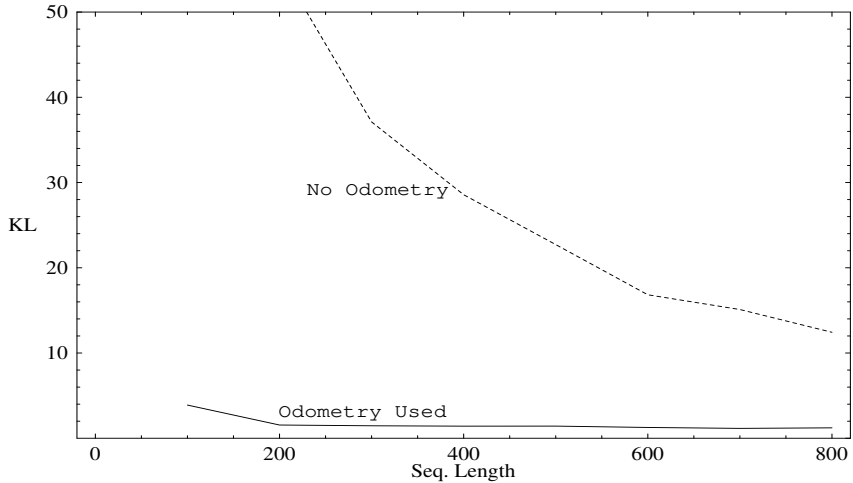

Figure 21: Average KL divergence as a function of sequence length.

## 7 Conclusions

Odometric information, which is often readily available in the robotics domain, makes it possible to learn hidden Markov models efficiently and effectively, while using shorter training sequences. More importantly, in contrast to the traditional perception of viewing the topological and the geometric models as two distinct types of entities, we have shown that the odometric information can be *directly* incorporated into the traditional topological HMM model, while maintaining convergence of the reestimation algorithm to a local maximum of the likelihood function.

Our method uses the odometric information in two ways. We first choose an initial model, based on the odometric information. An iterative procedure, which extends the Baum-Welch algorithm, is then used to learn the topological model of the environment while learning an additional set of constrained geometric parameters. The additional set of constrained parameters constitutes an extension to the basic HMM/POMDP model of transitions and observations. Even though we are primarily interested in the underlying topological model (transition and observation probabilities), our experiments demonstrate that the use of odometric relations can reduce the number of iterations and the amount of data required by the algorithm, and improve the resulting model.

The initialization procedure and the enforcement of the additivity constraint over relatively small models prove helpful both topologically and geometrically. An extensive study (Shatkay, 1999) shows that for long data sequences, generated from large models, enforcing only *anti-symmetry* rather than *additivity*, leads to better topological models. This is because in these cases, initialization is not always good, and additivity may over-constrain the learning to an unfavorable area. Learning large models may benefit from enforcing only anti-symmetry during the first few iterations, and complete additivity in later iterations. Alternatively, we may use our algorithm, enforcing additivity, to learn separate models for small portions of the environment, combining them later into one complete model. A similar idea of combining small model-fragments into a complete map of an environments was applied, in the context of geometrical maps, in recent work by Leonard and Feder (2000).

The work presented here demonstrates how domain-specific information and constraints can be enforced as part of the statistical estimation process, resulting in better models, while requiring shorter data sequences. We strongly believe that this idea can be applied in domains other than robotics. In particular, the acquisition of HMMs for use in molecular biology may greatly benefit from exploiting geometrical (and other) constraints on molecular structures. Similarly, temporal constraints may be exploited in domains in which POMDPs are appropriate for decision-support, such as air-traffic control and medicine.

## Acknowledgments

We thank Sebastian Thrun for his insightful comments throughout this work, John Hughes and Luis Ortiz for their helpful advice, Anthony Cassandra for his code for generating random distributions, Bill Smart for sustaining Ramona and Jim Kurien for providing the low level code for driving her. The presentation in this paper has benefited from the comments made by the anonymous referees to whom we are grateful. This work was done while both authors were at the Computer Science department at Brown University, and was supported by DARPA/Rome Labs Planning Initiative grant F30602-95-1-0020, by NSF grants IRI-9453383 and IRI-9312395, and by the Brown University Graduate Research Fellowship.

## Appendix A. An Overview of the Odometric Learning Algorithm

The algorithm takes as input an experience sequence $E = \langle r, V \rangle$, consisting of the odometric sequence $r$ and the observation sequence $V$, as defined in the beginning of Section 4.2. The number of states is also assumed to be given.

Learn Odometric HMM($E$)

| | | |
|---|---|---|
| 1 | Initialize matrices $A, B, R$ | (See Section 5) |
| 2 | $max\_change \leftarrow \infty$ | |
| 3 | **while** ( $max\_change > \epsilon$) | |
| 4 | **do** Calculate Forward probabilities, $\alpha$ | (Equation 4) |
| 5 | Calculate Backward probabilities, $\beta$ | (Equation 5) |
| 6 | Calculate state-occupation probabilities, $\gamma$ | (Equation 6) |
| 7 | Calculate State-transition probabilities, $\xi$, | (Equation 7) |
| 8 | $Old\_A \leftarrow A, \quad Old\_B \leftarrow B$ | |
| 9 | $A \leftarrow$ Reestimate ($A$) | (Equation 8, left) |
| 10 | $B \leftarrow$ Reestimate ($B$) | (Equation 8, right) |
| 11 | $R^\theta \leftarrow$ Reestimate ($R^\theta$) | (Equations 12 and 13) |
| 12 | $\langle R^x, R^y \rangle \leftarrow$ Reestimate($R^x, R^y$) | (Equations 10 and 11) |
| 13 | $max\_change \leftarrow$ MAX(Get_Max_Change($A, \ Old\_A$ ), Get_Max_Change($B, \ Old\_B$ )) | |

The equations referenced in Step 12 correspond to updates under the perpendicularity assumption, where a *global framework* is used. See (Shatkay, 1999) for update formulae within a state-relative framework.

If additivity is enforced, step 11 is followed by a projection of the reestimated $R^\theta$ onto an additive affine space, as described in Section 4.3.2. In addition, step 12 is substituted by the procedure described in Section 4.3.1. The reader is referred again to (Shatkay, 1999) for further detail.

Get_Max_Change is a function that takes two matrices and returns the maximal element-wise absolute difference between them. $\epsilon$ is a constant set to denote the margin of error on changes in parameters. When the change in parameters is "small enough", the model is regarded as "unchanged".